\documentclass[10pt,journal,compsoc]{IEEEtran}

\ifCLASSOPTIONcompsoc
  \usepackage[nocompress]{cite}
\else
  \usepackage{cite}
\fi

\usepackage{amsmath}

\usepackage{graphicx}
\usepackage{amssymb}
\usepackage{subcaption}
\usepackage{pifont}
\usepackage{placeins}
\usepackage{booktabs}
\usepackage{adjustbox}
\usepackage{booktabs}
\usepackage{multirow}
\usepackage{epigraph}
\usepackage{enumitem}
\usepackage{caption}
\usepackage{bbm, dsfont}
\usepackage{float}

\newcommand{\cmark}{\ding{51}}%
\newcommand{\xmark}{\ding{55}}%

\usepackage[usenames, dvipsnames]{color}
\usepackage{times}
\usepackage{epsfig}
\usepackage{graphicx}
\usepackage{float}
\usepackage{wrapfig}
\usepackage{amsmath,amssymb,amsthm}
\usepackage{algorithm,algorithmicx,algpseudocode}
\usepackage{bm,xspace}
\usepackage{comment}
\usepackage{verbatim}
\usepackage{multirow}
\usepackage{balance}
\usepackage{url}
\usepackage{booktabs}
\usepackage{etoolbox,siunitx}
\usepackage{calc}
\usepackage{pifont,hologo}
\usepackage{nicefrac}

\definecolor{TableauOrange}{RGB}{242,142,44}
\definecolor{TableauGreen}{RGB}{89,161,79}

\usepackage[super]{nth}
\usepackage{nicefrac}
\robustify\bfseries

\usepackage{dsfont}

\def\tt{\mathbf{t}}

\DeclareMathSymbol{@}{\mathord}{letters}{"3B}

\newcommand\mypara[1]{\vspace{1mm}\noindent\textbf{#1}}

\def\latex/{\LaTeX}
\def\bibtex/{\hologo{BibTeX}}

\usepackage{longtable,lscape}
\usepackage{rotating}

\newcommand{\red}[1]{{\color{red} {#1}}}
\newcommand{\blue}[1]{{\color{blue} {#1}}}

\usepackage{xcolor}
\definecolor{thing}{RGB}{255,0,0}
\definecolor{stuff}{RGB}{0,0,255}

\begin{document}
\title{MSeg: A Composite Dataset for Multi-domain Semantic Segmentation}


\author{John Lambert$^*$\thanks{$\;^*$Authors contributed equally.},
        Zhuang Liu$^*$,
        Ozan Sener,
        James Hays,
        and~Vladlen~Koltun% <-this % stops a space
\IEEEcompsocitemizethanks{
% \IEEEcompsocthanksitem 
% J. Lambert and Z. Liu contributed equally. 
\IEEEcompsocthanksitem J. Lambert and J. Hays are with the School of Interactive Computing, Georgia Institute of Technology.\protect
\IEEEcompsocthanksitem Z. Liu is with the Department of EECS, University of California, Berkeley.\protect
\IEEEcompsocthanksitem O. Sener and V. Koltun are with Intel Labs.}% <-this % stops an unwanted space
\thanks{}}%Manuscript received March, 2021; revised November, 2021.}}


\markboth{} %Submitted to TPAMI, March~2021}%
{Shell \MakeLowercase{\textit{et al.}}: Bare Demo of IEEEtran.cls for Computer Society Journals}



\IEEEtitleabstractindextext{%
\begin{abstract}
We present MSeg, a composite dataset that unifies semantic segmentation datasets from different domains. A naive merge of the constituent datasets yields poor performance due to inconsistent taxonomies and annotation practices. We reconcile the taxonomies and bring the pixel-level annotations into alignment by relabeling more than 220,000 object masks in more than 80,000 images, requiring more than 1.34 years of collective annotator effort. The resulting composite dataset enables training a single semantic segmentation model that functions effectively across domains and generalizes to datasets that were not seen during training. We adopt zero-shot cross-dataset transfer as a benchmark to systematically evaluate a model's robustness and show that MSeg training yields substantially more robust models in comparison to training on individual datasets or naive mixing of datasets without the presented contributions. A model trained on MSeg ranks first on the WildDash-v1 leaderboard for robust semantic segmentation, with no exposure to WildDash data during training. We evaluate our models in the 2020 Robust Vision Challenge (RVC) as an extreme generalization experiment. MSeg training sets include only three of the seven datasets in the RVC; more importantly, the evaluation taxonomy of RVC is different and more detailed. Surprisingly, our model shows competitive performance and ranks second. To evaluate how close we are to the grand aim of robust, efficient, and complete scene understanding, we go beyond semantic segmentation by training instance segmentation and panoptic segmentation models using our dataset. Moreover, we also evaluate various engineering design decisions and metrics, including resolution and computational efficiency. Although our models are far from this grand aim, our comprehensive evaluation is crucial for progress. Moreover, we share all the models and code with the community.
\end{abstract}

}

\maketitle

\IEEEdisplaynontitleabstractindextext

% \fi
\IEEEpeerreviewmaketitle

\IEEEraisesectionheading{\section{Introduction}\label{sec:introduction}}

%\section{Introduction}
\begin{figure*}
\centering
\begin{tabular}{@{}c@{}}
\includegraphics[width=\linewidth]{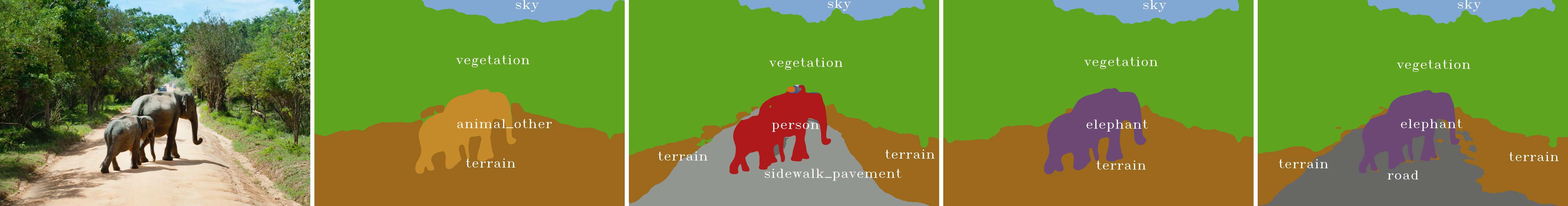} \\
\includegraphics[width=\linewidth]{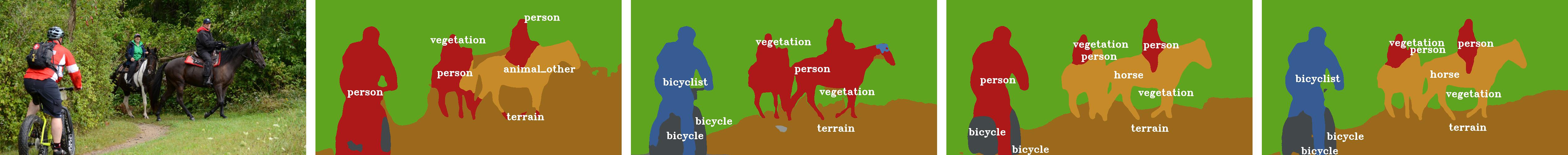} \\
\includegraphics[width=\linewidth]{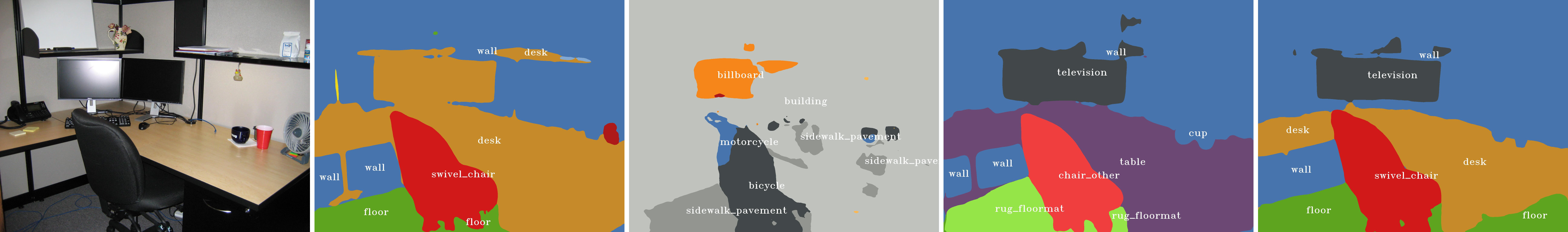}\\
\end{tabular}
  \begin{tabular}{@{}p{0.194\linewidth}@{\hspace{1mm}}p{0.194\linewidth}@{\hspace{1mm}}p{0.194\linewidth}@{\hspace{1mm}}p{0.194\linewidth}@{\hspace{1mm}}p{0.194\linewidth}@{}}
        \multicolumn{1}{@{}c@{}}{Input image} & 
          \multicolumn{1}{@{}c@{}}{ADE20K model} & \multicolumn{1}{@{}c@{}}{Mapillary model} &  
            \multicolumn{1}{@{}c@{}}{COCO model} &   \multicolumn{1}{@{}c@{}}{MSeg model}\\ &&&&
 \end{tabular}
    \caption{MSeg unifies multiple semantic segmentation datasets by reconciling their taxonomies and resolving incompatible annotations. This enables training models that perform consistently across domains and generalize better. Input images in this figure were taken from datasets that were not seen during training.}
           \label{fig:teaser}
\end{figure*}

When Papert first proposed computer vision as a summer project in 1966~\cite{Papert1966}, he described the primary objective as ``...a system of programs which will divide a vidisector picture into regions such as likely objects, likely background areas and chaos.'' Five decades later, computer vision is a thriving engineering field, and the task described by Papert is known as semantic segmentation~\cite{Shotton2009,Krahenbuhl11nips_CRFs, Long15cvpr_FCN,chen2015semantic,yu2016multi,Zhao:2017CVPR:PSPNet}.

Have we delivered on Papert's objective? A cursory examination of the literature would suggest that we have. Hundreds of papers are published every year that report ever-higher accuracy on semantic segmentation benchmarks such as Cityscapes~\cite{Cordts16cvpr_Cityscapes}, Mapillary~\cite{Neuhold:ICCV17:MapillaryDataset}, COCO~\cite{Lin:ECCV14:COCO}, ADE20K~\cite{Zhou:IJCV19:ADE20K}, and others. Yet a simple exercise can show that the mission has not been accomplished. Take a camera and begin recording as you traverse a sequence of environments: for example, going about your house to pack some supplies, getting into the car, driving through your city to a forest on the outskirts, and going on a hike. Now perform semantic segmentation on the recorded video. Is there a model that will successfully perform this task?

A computer vision professional will likely resort to multiple models, each trained on a different dataset. Perhaps a model trained on the NYU dataset for the indoor portion~\cite{Silberman:ECCV12:nyudepthv2}, a model trained on Mapillary for the driving portion, and a model trained on ADE20K for the hike. Yet this is not a satisfactory state of affairs. It burdens practitioners with developing multiple models and implementing a controller that decides which model should be used at any given time. It also indicates that we haven't yet arrived at a satisfactory vision system: after all, an animal can traverse the same environments with a single visual apparatus that continues to perform its perceptual duties throughout.

%In hindsight, practitioners still require such a model which, given any image, can segment it into a set of semantic objects. In this work, we look at this problem from a practitioners' perspective and ask \emph{do we have the right set of tools, datasets and algorithms, to give practitioners a practical semantic segmentation software?}

A natural solution is to train a model on multiple datasets, hoping that the result will perform as well as the best dedicated model in any given environment. As has previously been observed, and confirmed in our experiments, the results are far from satisfactory. A key underlying issue is that different datasets have different taxonomies: that is, they have different definitions of what constitutes a `category' or `class' of visual entities. Taxonomic clashes and inconsistent annotation practices across datasets from different domains (e.g., indoor and outdoor, urban and natural, domain-specific and domain-agnostic) substantially reduce the accuracy of models trained on multiple datasets.

In this paper, we take steps towards addressing these issues. We present MSeg, a composite dataset that unifies semantic segmentation datasets from different domains: COCO~\cite{Lin:ECCV14:COCO}, ADE20K~\cite{Zhou:IJCV19:ADE20K}, Mapillary~\cite{Neuhold:ICCV17:MapillaryDataset}, IDD~\cite{Varma:WACV19:IDD}, BDD~\cite{Yu:Arxiv2018:BDD}, Cityscapes~\cite{Cordts16cvpr_Cityscapes}, and SUN RGB-D~\cite{Song:2015:CVPR:SUN-RGBD}. A naive merge of the taxonomies of the seven datasets would yield more than 300 classes, with substantial internal inconsistency in definitions. Instead, we reconcile the taxonomies, merging and splitting classes to arrive at a unified taxonomy with 194 categories. To bring the pixel-level annotations in conformance with the unified taxonomy, we conduct a large-scale annotation via the Mechanical Turk platform and produce compatible annotations across datasets by relabeling object masks.

The resulting composite dataset enables training unified semantic segmentation models that come to a step closer to delivering on Papert's vision. MSeg training yields models that exhibit much better generalization to datasets that were not seen during training. We adopt \emph{zero-shot cross-dataset transfer} as a proxy for a model's expected performance in the ``real world''~\cite{Lasinger:2019:RobustDepthMix}. We train models on MSeg and test on datasets that are disjoint from MSeg. In this mode, MSeg training is substantially more robust than training on individual datasets or training on multiple datasets without the reported taxonomic reconciliation.

Accurate semantic segmentation is typically not enough for a typical computer vision practitioner as it lacks any sense of an object. Instance segmentation extends semantic segmentation with finding semantically meaningful instances, and panoptic segmentation \cite{Kirillov19cvpr_PanopticSeg} combines instance segmentation with semantic segmentation. Similarly, we evaluate the capabilities of MSeg dataset for these problems as the component training datasets have labels for the instances. Moreover, the practical applications typically come with computational efficiency requirements. For this purpose, we further evaluate the computational efficiency of our models at various resolutions to understand the computational efficiency and accuracy trade-offs. Although the results of these experiments are still far from the ultimate requirements of the practitioners, it gives a complete understanding of the MSeg dataset and models. Moreover, we share all trained models, data, and source code with the community.

\section{Related Work}

\subsection{Cross-domain semantic segmentation}
Annotation for semantic segmentation is an exceptionally costly process as pixel-level annotations are needed. Utilizing multiple or additional existing semantic segmentation datasets is a sensible direction to ease this cost and effectively increase the dataset size. Mixing segmentation datasets has primarily been done within a single domain and application, such as driving. Ros et al.~\cite{Ros:Arxiv16:ConstrainedDeconvRoad} aggregated six driving datasets to create the MDRS3 meta-dataset, with approximately 30K images in total. Bevandic et al.~\cite{Bevandic:GCPR2019:SemSegOutlierDet} mix Mapillary Vistas, Cityscapes, the WildDash validation set, and \mbox{ImageNet-1K-BB} (a subset of ImageNet \cite{Deng:cvpr09:imagenet} for which bounding box annotations are available) for joint segmentation and outlier detection on WildDash \cite{Zendel:2018:ECCV:WildDash}. On a smaller scale, \cite{Meletis:IVS2018:heterogeneousdatasets, leonardi:iciap2019:multipledatasets} mix Mapillary, Cityscapes, and the German Traffic Sign Detection Benchmark. In contrast to these works, we focus on semantic segmentation across multiple domains and resolve inconsistencies between datasets at a deeper level, including relabeling incompatible annotations.

Varma et al.~\cite{Varma:WACV19:IDD} evaluate the transfer performance of semantic segmentation datasets for driving. They only use 16 common classes, without any dataset mixing. They observe that cross-dataset transfer is significantly inferior to ``self-training'' (i.e., training on the target dataset). We have observed the same outcomes when models are trained on individual datasets, or when datasets are mixed naively. 

Liang et al.~\cite{Liang:2018:CVPR:DSSPN_Universal} train a model by mixing Cityscapes, ADE20K, COCO Stuff, and Mapillary, but do not evaluate cross-dataset generalization. Kalluri et al.~\cite{Kalluri:2019:ICCV:UniversalSemiSupervised} mix pairs of datasets (Cityscapes + CamVid, Cityscapes + IDD, Cityscapes + SUN RGB-D) for semi-supervised learning.

An underlying issue that impedes progress on unified semantic segmentation is the incompatibility of dataset taxonomies. In contrast to the aforementioned attempts, we directly address this issue by deriving a consistent taxonomy that bridges datasets from multiple domains.

\subsection{Domain adaptation and generalization}
Training datasets are biased and deployment in the real world presents the trained models with data that is unlike what had been seen during training ~\cite{Torralba:2011:DatasetBias}. This is known as \textit{covariate shift} \cite{Shimodaira2000} or \textit{selection bias} \cite{Heckman1979}, and can be tackled in the \emph{adaptation} or the \emph{generalization} setting. In \emph{adaptation}, samples from the test distribution (deployment environment) are available during training, albeit without labels. In \emph{generalization}, we expect models to generalize to previously unseen environments after being trained on data from multiple domains. An extensive collection of these methods have recently been summarized and benchmarked in various settings \cite{Gulrajani21iclr_lostdg, Koh21icml_wilds}, and we refer the interested reader to these studies. We summarize the most relevant ones and discuss their relationship with our work.

We operate in the generalization mode and aim to train robust models that perform well in new environments, with no data from the target domain available during training. Many domain generalization approaches are based on the assumption that learning features that are invariant to the training domain will facilitate generalization to new domains~\cite{Motiian:2017:ICCV:CCSA,Mancini2018}. Volpi et al.~\cite{Volpi:NIPS2018:AdversarialAug} use distributionally robust optimization by considering domain difference as noise in the data distribution space. Bilen and Vedaldi~\cite{Bilen:Arxiv17:UniversalRep} propose to learn a unified representation and eliminate domain-specific scaling factors using instance normalization. Mancini et al.~\cite{Mancini2018} modify batch normalization statistics to make features and activations domain-invariant. Recently, the tools from causality has been applied to the problem of domain generalization~\cite{Arjovsky2019_IRM, Chen2020_Causal}. The causal relationship between images and their labels are expected to generalize to novel domains whereas environment specific relationships are expected to be anti-causal.

The aforementioned domain generalization methods assume that the same classifier can be applied in all environments. This relies on compatible definitions of visual categories which our work enables. Our work is complementary and can facilitate future research on domain generalization by providing a compatible taxonomy and consistent annotations across semantic segmentation datasets from different domains.

\subsection{Visual learning over diverse domains}
In order to study the problem of learning from multiple datasets, the Visual Domain Decathlon \cite{Rebuffi:NIPS2017:ResidualAdapter} introduced a benchmark over ten image classification datasets. However, it allows training on all of them. More importantly, its purpose is not training a single classifier. Instead, they hope domains will assist each other by transferring inductive biases in a multi-task setting. Triantafillou et al.~\cite{Triantafillou:Arxiv2019:Metadataset} proposed a meta-dataset for benchmarking few-shot classification algorithms.

For the problem of monocular depth estimation, Ranftl et al.~\cite{Lasinger:2019:RobustDepthMix} use multiple datasets and mix them via a multi-task learning framework. We are inspired by this work and aim to facilitate progress on dataset mixing and cross-dataset generalization in semantic segmentation. Unlike the work of Ranftl et al., which dealt with a geometric task (depth estimation), we are confronted with inconsistencies in semantic labeling across datasets, and make contributions towards resolving these.

A recent robust vision challenge \cite{RVC} addresses robustness and generalization by learning from multiple datasets. The challenge is using our dataset but in a \emph{few-shot generalization} setting instead of \emph{zero-shot generalization}. Although our models are trained in the \emph{zero-shot} regime, the resulting performance is competitive with \emph{few-shot} models showing our approach is a promising direction to develop general and robust vision models. 

\subsection{Beyond Semantic Segmentation}
Semantic segmentation divides an image into meaningful sub-regions; however, it fails to provide any information about instance and object properties. For example, a large crowd is segmented as people regardless of the number of people within the region. In order to overcome this limitation, object detection and segmentation have been jointly studied under the term of image parsing \cite{Tu05IJCV_Image_Parsing}. This problem has further been studied using graphical models \cite{Tighe13cvpr_Finding_Things, Tighe14cvpr_Scene_Parsing, Yao12cvpr_Describing}. 

Recently the problem of instance segmentation has gained interest. It segments an image into semantically meaningful instances \cite{Arnab17cvpr,Bai17cvpr, Liu17iccv_SGN, Kirillov17cvpr_InstanceCut, Liang20cvpr_PolyTransform}. However, it only consider \emph{things}, not \emph{stuff} \cite{Adelson91book_PlenopticFn}. Panoptic segmentation \cite{Kirillov19cvpr_PanopticSeg} is an emerging scene understanding task which jointly performs instance segmentation and semantic segmentation over both thing and stuff classes. 

The vast majority of train datasets in our composite dataset include instance labels; hence, we extend our dataset to instance segmentation and panoptic segmentation. We show successful panoptic and instance segmentation results using our dataset, validating the usefulness of MSeg for these future directions.

\section{The MSeg Dataset}
%\subsection{Component Datasets}
%\label{sec:dataset_composition}

Our proposed dataset, MSeg, is a composite dataset as it is composed of multiple diverse datasets after a label unification process. We consider a large number of semantic segmentation datasets and select a subset to create MSeg. We list the semantic segmentation datasets used in MSeg in Table~\ref{tab:existingdatasets}. The datasets that were not used, and reasons for not including them, are listed in the Table~\ref{tab:excludeddatasets}.

\begin{table}[t]
\centering
\caption{Component datasets in MSeg.}
\begin{adjustbox}{max width=\columnwidth}
\begingroup
\renewcommand{\arraystretch}{1.25} % General space between rows (1 standard)
    \begin{tabular}{@{}p{1mm}@{\hspace{4mm}}l@{\hspace{4mm}}l@{\hspace{4mm}}r@{}}
    \toprule 
     \multicolumn{2}{@{}l@{}}{\textbf{Dataset name}} &  \textbf{Origin domain} &  \textbf{\# Images}  \\
    \midrule
    \multicolumn{3}{@{}l@{}}{\textbf{Training \& Validation}} & \\
       &   \textsc{COCO} \cite{Lin:ECCV14:COCO}
        \textsc{+ COCO Stuff} \cite{Caesar:CVPR18:COCOStuff}
          &  Everyday objects & 123,287 \\
      & ADE20K \cite{Zhou:IJCV19:ADE20K} &  Everyday objects & 22,210 \\
      &  \textsc{Mapillary} \cite{Neuhold:ICCV17:MapillaryDataset} & Driving (Worldwide) & 20,000 \\
      &  \textsc{IDD} \cite{Varma:WACV19:IDD} &  Driving (India) &  7,974 \\
      & BDD \cite{Yu:Arxiv2018:BDD} &  Driving (United States) & 8,000 \\
      &  \textsc{Cityscapes} \cite{Cordts16cvpr_Cityscapes} &  Driving (Germany) & 3,475 \\
      &  \textsc{SUN RGBD}  \cite{Song:2015:CVPR:SUN-RGBD} &  Indoor & 5,285 \\
    \multicolumn{3}{@{}l@{}}{\textbf{Test}} & \\
       &  \textsc{PASCAL VOC} \cite{Everingham10ijcv_PASCAL} &  Everyday objects &  1,449\\
       & \textsc{PASCAL Context} \cite{Mottaghi:cvpr14:context} & Everyday objects & 5,105 \\
       &   \textsc{CamVid} \cite{Brostow:2009:CamVid} &  Driving (U.K.) & 101 \\
       &  \textsc{WildDash-v1}\footnotemark  \cite{Zendel:2018:ECCV:WildDash} &  Driving (Worldwide) & 70 \\
       & \textsc{KITTI} \cite{Geiger:2013IJRR:Kitti} & Driving (Germany) & 200 \\
       & \textsc{ScanNet-20} \cite{Dai:2017CVPR:ScanNet} & Indoor  & 5,436 \\
        %  &   SUN RGB-D \cite{Song:2015:CVPR:SUN-RGBD} &  Indoor & \\
         \bottomrule 
    \end{tabular}
\endgroup
\end{adjustbox}
\label{tab:existingdatasets}
\end{table}

Our guiding principle for selecting a training/test dataset split is that large, modern datasets are most useful for training, whereas older and smaller datasets are good candidates for testing. 
We test zero-shot cross-dataset performance on the validation subsets of these datasets. Note that the data from the test datasets (including their training splits) is never used for training in MSeg. For validation, we use the validation subsets of the training datasets listed in Table~\ref{tab:existingdatasets}. Next, we describe the dataset-specific choices we make.

\begin{table*}[t]
        \caption{List of datasets we do not currently include in MSeg. }
\begin{adjustbox}{max width=\textwidth}
\begingroup
\renewcommand{\arraystretch}{1.25} % General space between rows (1 standard)
    \begin{tabular}{@{}p{1mm}@{\hspace{4mm}}l@{\hspace{4mm}}l@{\hspace{4mm}}r@{\hspace{8mm}}p{8cm}@{}}
    \toprule 
     \multicolumn{2}{@{}l@{}}{\textbf{Dataset Name}} &  \textbf{Origin Domain} &  \textbf{\# Images}  & \textbf{Reason for Exclusion} \\
    \midrule
    \multicolumn{3}{@{}l@{}}{\textbf{Excluded Datasets}} & \\
      &  \textsc{ApolloScape} \cite{Huang:2018CVPRW:ApolloScape} & Driving (Worldwide) & 147,000 & We include only fully-manually labeled datasets in MSeg. ApolloScape uses a semi-automatic label generation pipeline. \\
      &  \textsc{VIPER} \cite{Richter:2016:ECCV:PlayForData} &  Driving (Synthetic) & 134,097 & We do not use synthetic data. \\
      &  \textsc{InteriorNet} \cite{Li:BMVC18:InteriorNet} &  Indoor (Synthetic) & 20,000,000  & We do not use synthetic data. \\
      &  \textsc{Matterport3D} \cite{Chang:20173DV:Matterport} &  Indoor &  194,400  & Annotated in 3D, rather than in 2D. \\
      &  \textsc{Stanford 2D-3D-S} \cite{Armeni:Arxiv2017:2D-3D-S} &  Indoor  & 70,496  & Data is semantically annotated on 3D point clouds, rather than in 2D; the dataset includes only 13 object classes, which is few for an indoor environment. \\
      & \textsc{OpenImages v5} \cite{Benenson19cvpr_OIDv5,Kuznetsova20ijcv_OpenImages,Krasin17_OpenImages2} & Everyday Objects & 957,561 & Provides only object segmentations (no segmentation of stuff categories). \\
       &  \textsc{A2D2} \cite{Geyer:2019:A2D2}  &  Driving (Germany) &  40,000 & Was not available at time of paper preparation. \\
       &  \textsc{MINC OpenSurfaces} \cite{Bell15cvpr_MINC}   & Materials  & & Annotation categories are material-centric, instead of object- or stuff-centric.  \\
       &  \textsc{Replica Dataset} \cite{Straub:2019:Replica}  & Indoor (Synthetic) & N/A & Real data is not available (images are synthetically rendered).  \\
         \bottomrule
    \end{tabular}
\endgroup
\end{adjustbox}
    \label{tab:excludeddatasets}
\end{table*}

We use the free, academic version of Mapillary Vistas~\cite{Neuhold:ICCV17:MapillaryDataset}. In this we forego highly detailed classification of traffic signs, traffic lights, and lane markings in favor of broader access to MSeg.

For COCO \cite{Lin:ECCV14:COCO}, we use the taxonomy of COCO Panoptic as a starting point, rather than COCO Stuff~\cite{Caesar:CVPR18:COCOStuff}. The COCO Panoptic taxonomy merges some of the material-based classes of COCO Stuff into common categories that are more compatible with other datasets. (E.g., \textit{floor-marble}, \textit{floor-other}, and \textit{floor-tile} are merged into \textit{floor}.)

Naively combining the component datasets yields roughly 200K images with 316 semantic classes (after merging classes with synonymous names). We found that training on naively combined datasets yields low accuracy and poor generalization. We believe the main cause for this failure is inconsistency in the taxonomies and annotation practices in the different datasets. The following subsections explain these issues and our solution.

\footnotetext{A second version of the benchmark, \emph{WildDash-v2}, was introduced after the completion of our experiments and publication of our CVPR 2020 paper.}

\subsection{Taxonomy}
\label{sec:unified-tax-principles}

In order to train a cross-domain semantic segmentation model, we need a unified taxonomy. In order to unify the training taxonomies, we need to perform merging and splitting operations. Merging and splitting of classes from component datasets have different drawbacks. Merging is easy and can be performed programmatically, with no additional labeling. The disadvantage is that labeling effort that was invested into the original dataset is sacrificed and the resulting taxonomy has coarser granularity. Splitting, on the other hand, is labor-intensive. To split a class from a component dataset, all masks with that class need to be relabeled. This provides finer granularity for the resulting taxonomy, but costs time and labor.

\begin{figure}[t]
    \centering
    \includegraphics[width=\columnwidth]{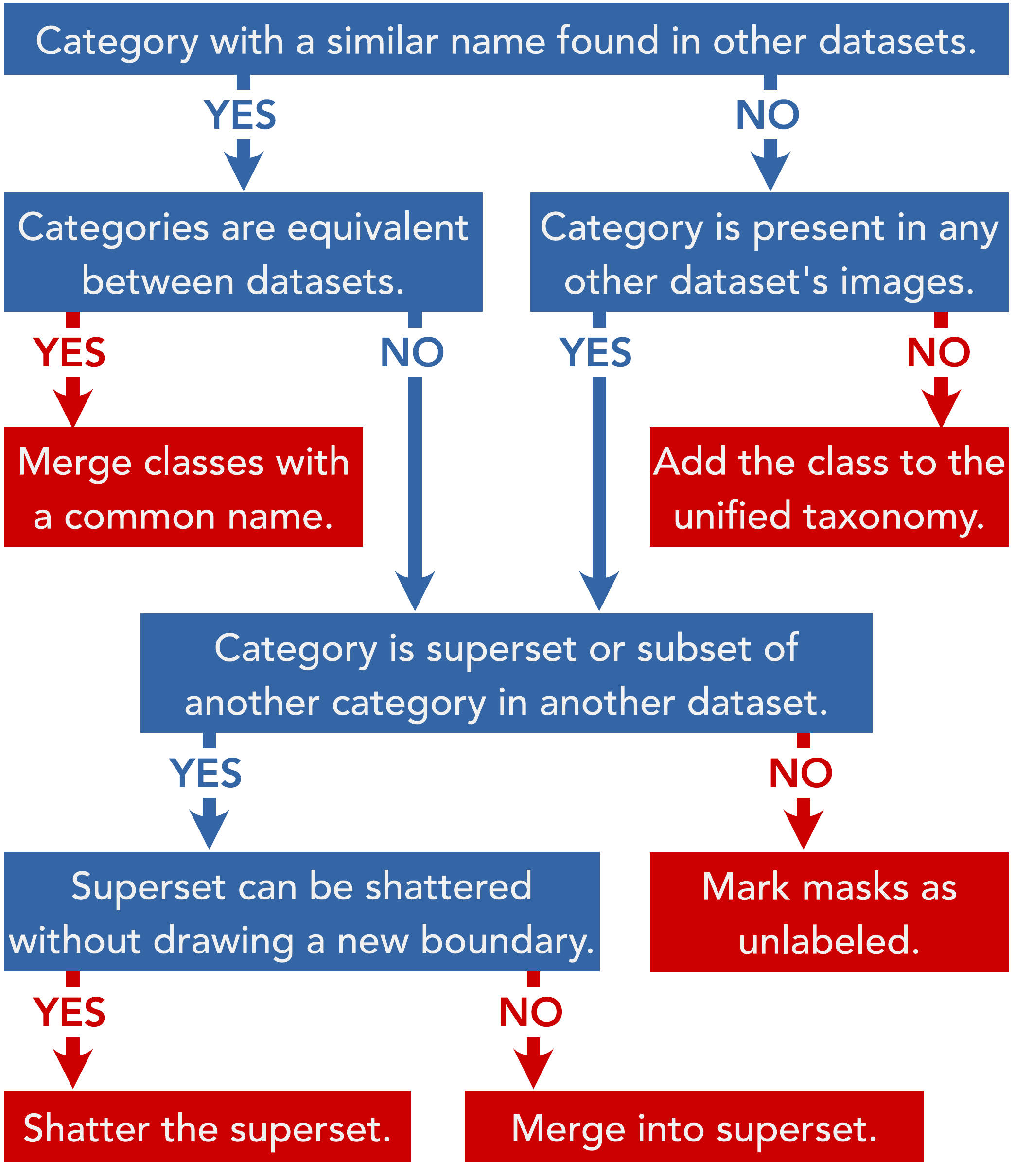}
    \caption{Procedure for determining the set of categories in the MSeg taxonomy.}
    \label{fig:decisiontree}
\end{figure}

In order to make these decisions, we consider two primary objectives. First, as many classes should be preserved as possible. %For example, \textit{guardrail} should not be discarded just because COCO, BDD, or IDD do not annotate it.
Merging classes can reduce the discriminative ability of the resulting models. Second, the taxonomy should be flat, rather than hierarchical, to maximize compatibility with standard training methods. In order to satisfy these objectives, we followed a sequence of decision rules, summarized in Figure~\ref{fig:decisiontree}, to determine split and merge operations on taxonomies of the component datasets.  Although the choice of classes in any taxonomy is somewhat arbitrary (e.g. a taxonomy for a fashion dataset versus a wildlife preservation dataset will be highly-task dependent), our decision tree allows us to make objective choices. 

\begin{figure}
    \centering
    \includegraphics[width=0.9\columnwidth]{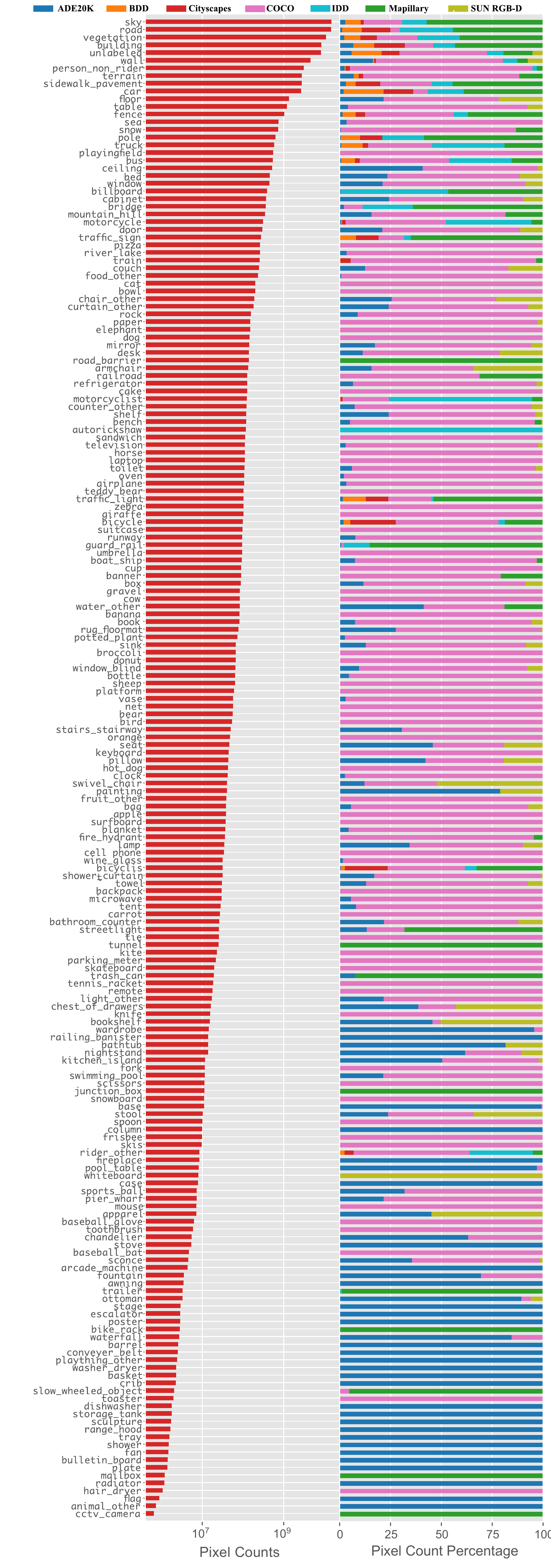}
    \caption{Semantic classes in MSeg. Left: pixel counts of MSeg classes, in log scale. Right: percentage of pixels from each component dataset that contribute to each class. Any single dataset is insufficient for describing the visual world.}
    \label{fig:universal-dataset-class-contributions}
\end{figure}

Given a query category, if a category with a similar or related name is found in any one of the other datasets’ taxonomies, we manually inspect the image masks. If we determine the category meaning to be equivalent, (e.g. Mapillary ``On Rails'' and COCO-Panoptic ``train'''), we combine class masks under a common name.

If no similar or related name is found in any other dataset's taxonomy, we must inspect a large number of images from each such dataset in order to determine if the category is present, but simply unannotated or annotated differently. If a category is truly unique to this dataset (i.e. present in no other dataset), we add this class to the unified taxonomy (e.g. IDD's ``auto-rickshaw'' or COCO-Panoptic's ``surfboard'' or ``tie''). 

In the case that a similar category name exists, but was not equivalent (e.g. COCO-Panoptic ``person'' and Mapillary ``person''), or a category is present in another dataset's images, a more challenging task remains. In such a scenario, we seek to answer the question, \textit{is there a category in any other dataset that is a strict superset or subset of query category?} If superset/subsets were to be merged, current boundaries must keep unrelated classes separate. If the answer is no, (e.g. Mapillary ``manhole'' vs. COCO-Panoptic's ``road, or IDD's ``non-drivable fallback'' and Cityscapes ``terrain''), we mark masks as ``unlabeled.'' Given there is no hierarchical relationship, one would need to re-draw new boundaries (an endeavor which we do not undertake).

However, if there is a category in another dataset that is a strict superset or subset of the query category (e.g. ADE20K ``glass'' and COCO-Panoptic's ``wine glass''), there is a possibility to ``shatter'' the superset category. Shattering the superset must be enforceable to an MTurk worker, given a typical field of view, without drawing a new boundary (e.g.  ADE20K ``swivel-chair'' and COCO-Panoptic ``chair'' is possible since boundaries are shared, and the field of view generally accommodates this, unlike “building" vs. "house”). If some shattered masks do not fall into any pre-existing subset, we add an ``-other'' category. However, if such shattering is impossible without drawing a new boundary, (e.g. ADE20K’s ``flower'', ``tree'', ``palm'' vs. Cityscapes ``vegetation''), we merge both into the superset. If such shattering is impossible because the classification is not enforceable to a Mechanical Turk worker given a typical field-of-view (e.g. COCO-Panoptic’s ``dining table'' and ``table-merged''), we merge both into the superset category.

\begin{figure*}
    \centering
    \vspace{-5mm}
    \includegraphics[width=0.96\textwidth]{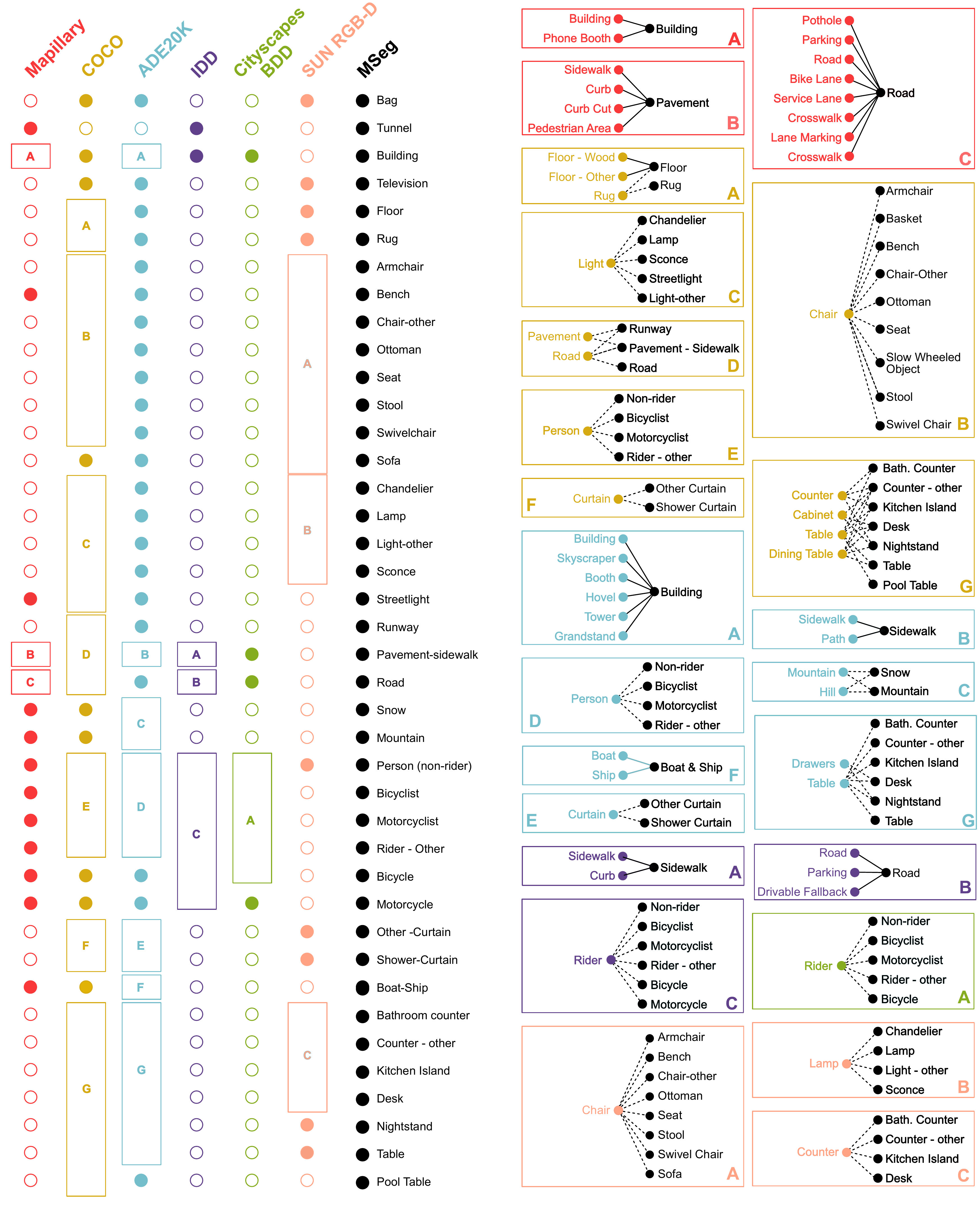}
    \vspace{-2mm}
    \caption{Visualization of a subset of the class mapping from each dataset to our unified taxonomy. This figure shows 40 of the 194 classes; see the appendix for the full list. Each filled circle means that a class with that name exists in the dataset, while an empty circle means that there is no pixel from that class in the dataset. A  rectangle indicates that a split and/or merge operation was performed to map to the specified class in MSeg. Rectangles are zoomed-in in the right panel. Merge operations are shown with straight lines and split operations are shown with dashed lines. \emph{(Best seen in color.)}}
    \label{fig:dataset}
\end{figure*}

Following this process, we condensed the 316 classes obtained by merging the component datasets into a unified taxonomy of 194 classes. The full list and class statistics are given in Figure~\ref{fig:universal-dataset-class-contributions} and further described and visualized in the appendix. Each of these classes is derived from classes in the component datasets. We also report the statistics of the number of relabeled masks in the appendix.

After this process, an MSeg category can have one of the following relationships to classes in a component dataset: (a) it can be in direct correspondence to a class in a component taxonomy, (b) it can be the result of merging a number of classes from a component taxonomy, (c) it can be the result of splitting a class in a component taxonomy (one-to-many mapping), or (d) it can be the union of classes which are split from different classes in the component taxonomy.

Figure~\ref{fig:dataset} visualizes these relationships for 40 classes. For example, the class `person' in COCO and ADE20K corresponds to four classes (`person', `rider-other', `bicyclist', and `motorcyclist') in the Mapillary dataset. Thus the `person' labels in COCO and ADE20K need to be split into one of the aforementioned four Mapillary categories depending on the context. (See boxes COCO-E and ADE20K-D in Figure~\ref{fig:dataset}.) Mapillary is much more fine-grained than other driving datasets and classifies \emph{Pothole, Parking, Road, Bike Lane, Service Lane, Crosswalk-Plain, Lane Marking-General, Lane Marking-Crosswalk} separately. These classes are merged into a unified MSeg `road' class. (See box Mapillary-C in Figure~\ref{fig:dataset}.)

%The next subsection focuses on the splitting operation and explains the relabeling process.
\subsection{Relabeling Instances of Split Classes}
\label{sec:amt}

We utilize Amazon Mechanical Turk (AMT) to relabel masks of classes that need to be split. We re-annotate only the datasets used for learning, leaving the evaluation datasets intact. Instead of recomputing boundaries, we formulate the problem as multi-way classification and ask annotators to classify each mask into finer-grained categories from the MSeg taxonomy. We include an example labeling screen, workflow and labeling validation process in the appendix  In total, we split 31 classes and relabel 221,323 masks. We visualize some of these split operations in Figure~\ref{fig:dataset} and provide additional details in the appendix.

\begin{table*}[t]
\begin{tabular}{@{}c@{}}
\includegraphics[width=\linewidth]{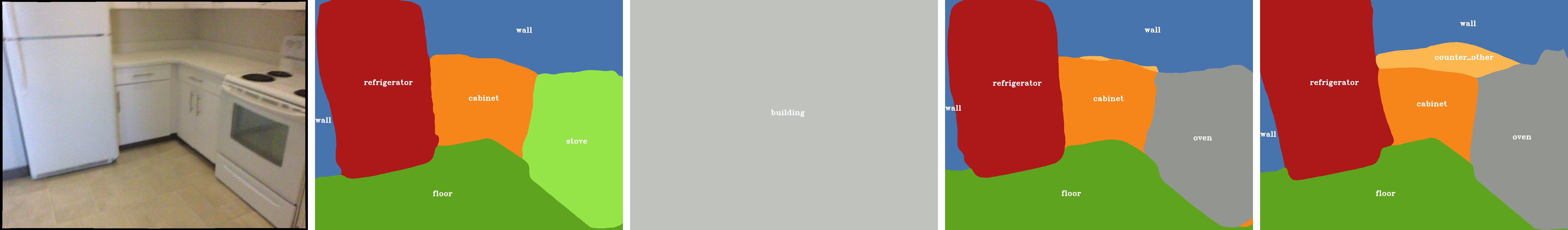}\\
\includegraphics[width=\linewidth]{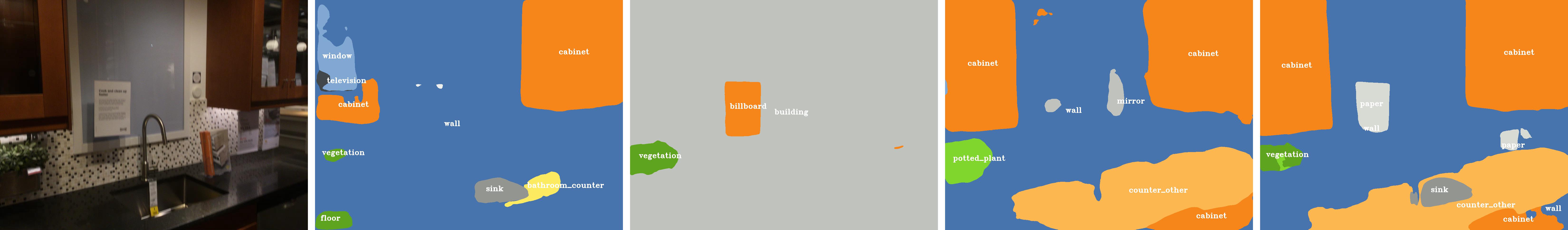}\\
\includegraphics[width=\linewidth]{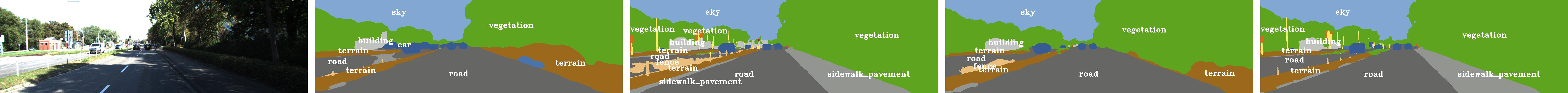}\\
\includegraphics[width=\linewidth]{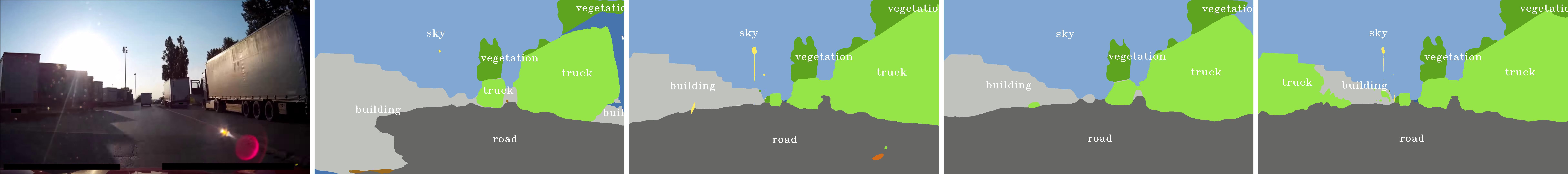}\\
\includegraphics[width=\linewidth]{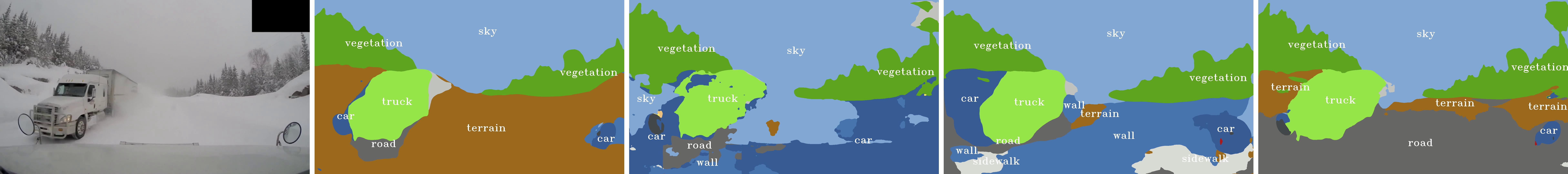}\\
\end{tabular}
  \begin{tabular}{@{}p{0.19\linewidth}@{\hspace{2mm}}p{0.19\linewidth}@{\hspace{2mm}}p{0.19\linewidth}@{\hspace{2mm}}p{0.19  \linewidth}@{\hspace{2mm}}p{0.19\linewidth}@{}}
        \multicolumn{1}{@{}c@{}}{Input image} &
          \multicolumn{1}{@{}c@{}}{ADE20K model} & \multicolumn{1}{@{}c@{}}{Mapillary model} &  
            \multicolumn{1}{@{}c@{}}{COCO model} &   \multicolumn{1}{@{}c@{}}{MSeg model}\\ &&&& \\
 \end{tabular}
 \vspace{-3mm}
 \captionof{figure}{Zero-shot generalization on images from MSeg test datasets; ScanNet-20 (top two rows), KITTI (middle row), and WildDash (bottom two rows). In the fourth row the objects on the left are semi-truck trailers, and only the MSeg model recognizes them.}
 \label{fig:qualexamples-eachdataset}
 \vspace{3mm}
    \captionof{table}{Semantic segmentation accuracy (mIoU) on MSeg test datasets using 1 million crops. (Zero-shot cross-dataset generalization.) \textit{Top:} performance of models trained on individual training datasets. \textit{Middle:} the same model trained on MSeg (our result). \textit{Bottom:} for reference, performance of `oracle' models trained on the test datasets. Numbers within 1\% of the best are in bold (excluding `oracle' models). The rightmost column is a summary measure: harmonic mean across datasets. \emph{Inference is performed at a single scale.} }
    \vspace{1mm}
    \centering
    \small
    %\resizebox{\columnwidth}{!}{
    % \begin{tabular}{@{}lccccccc@{}}
        \begin{tabular}{@{}lccccccc@{}}
   \toprule
       Train/Test & VOC & Context & CamVid & WildDash   & KITTI  & ScanNet & \textit{h.\ mean} \\
       \midrule
                  COCO  &  $  \mathbf{73.4} $ & $  \mathbf{43.3} $ & $  58.7 $ & $  38.2 $ & $  47.6 $ & $  33.4 $ & $  45.8 $\\
                ADE20K  &  $  35.4 $ & $  23.9 $ & $  52.6 $ & $  38.6 $ & $  41.6 $ & $  42.9 $ & $  36.9 $\\
             Mapillary  &  $  22.5 $ & $  13.6 $ & $  82.1 $ & $  55.4 $ & $  \mathbf{67.7} $ & $   2.1 $ & $   9.3 $\\
                   IDD  &  $  14.6 $ & $   6.5 $ & $  72.1 $ & $  41.2 $ & $  51.0 $ & $   1.6 $ & $   6.5 $\\
                   BDD  &  $  14.4 $ & $   7.1 $ & $  70.7 $ & $  52.2 $ & $  54.5 $ & $   1.4 $ & $   6.1 $\\
            Cityscapes  &  $  13.3 $ & $   6.8 $ & $  76.1 $ & $  30.1 $ & $  57.6 $ & $   1.7 $ & $   6.8 $\\
              SUN RGBD  &  $  10.0 $ & $   4.3 $ & $   0.1 $ & $   1.9 $ & $   1.1 $ & $  42.6 $ & $   0.3 $\\
\midrule
               MSeg  &  $  70.7 $ & $  \mathbf{42.7} $ & $  \mathbf{83.3} $ & $  \mathbf{62.0} $ & $  \mathbf{67.0} $ & $  \mathbf{48.2} $ & $  \mathbf{59.2} $\\
MSeg-w/o relabeling  &  $  70.2 $ & $  \mathbf{42.7} $ & $  82.0 $ & $  \mathbf{62.7} $ & $  65.5 $ & $  43.2 $ & $  57.6 $\\
\midrule
           Oracle &  $  77.8 $ & $  45.8 $ & $  78.8 $ & -- & $  58.4 $ & $  62.3 $ & --\\
       \bottomrule
        \vspace{-3mm}
    \end{tabular}
    %}
    \label{tab:resultsontestdatasets}
\centering
\end{table*}

AMT workers sometimes submit inaccurate, random, or even adversarial decisions~\cite{Snow2008}. To ensure annotation quality, we embed `sentinel' tasks within each batch of work~\cite{Chen17iccv_CascadedRefinement,Richter17iccv_PlayingForBenchmarks,Gupta19cvpr_LVIS}, constituting at least 10\% of each batch. These sentinels are tasks for which the ground truth is unambiguous and is manually annotated by the authors. We use the sentinels to automatically evaluate the reliability of each annotator so that we can direct work towards more reliable annotators. In order to guarantee acceptable accuracy, five workers annotate each batch, and the work is resubmitted until all submitted batches meet a 100\% sentinel accuracy. Afterwards, the category is determined by majority vote; categories that do not meet these criteria are manually labeled in-house by an expert annotator (one of the authors).

\section{Experimental Results}
%\subsection{Implementation}
%\label{sec:implementation}

We quantitatively and qualitatively study MSeg for learning multi-domain scene understanding models. We start with discussing the implementation details in Section~\ref{sec:implementation}. Then, we evaluate the performance of semantic segmentation models trained using MSeg data and the impact of our universal taxonomy in Section~\ref{sec:zeroshot_exp}. Finally, we study the practical multi-domain scene understanding problem beyond semantic segmentation. We train multi-domain semantic, instance, and panoptic segmentation models using larger compute and evaluate them in terms of accuracy and efficiency. We share the source code, trained models, and the dataset for full reproducibility\footnote{Trained models, data, and source code: \textcolor{blue}{\url{https://github.com/mseg-dataset}}}.

\subsection{Implementation details}
\label{sec:implementation}
Our semantic segmentation experiments largely follow the protocol introduced by Zhao et al.~\cite{ZhaoImplementation}. However, we used the HRNet-V2-W48 architecture~\cite{Sun19arxiv_HRNetSegmentation}, where W48 indicates the width of the high-resolution convolution.  We use HRNet without Object Contextual Representations (OCR) and with synchronized BN. 
For instance segmentation, we use Mask R-CNN \cite{He17iccv_MaskRCNN} with a ResNeXt-101-32x8d-FPN \cite{Xie17cvpr_ResNeXt} backbone. For panoptic segmentation, we use Panoptic-FPN \cite{Kirillov19cvpr_PanopticFPN} with a ResNet-101 backbone. We use 1080p resolution in our experiments with the crop size of $713 \times 713$. When we further evaluate the impact of resolution, we use crop sizes of $473 \times 473$ for 480p and $593 \times 593$ for 720p.

We use SGD with momentum 0.9 and a weight decay of $ 10^{-4}$. We use a polynomial learning rate with power 0.9. Base learning rate is set to $10^{-2}$.  For multi-scale inference, we use a multi-scale accumulation of probabilities, with scales of 0.5 to 1.75 with 0.25 increments. % An auxiliary cross-entropy loss is added to intermediate activations, a linear combination with weight 0.4.

When forming a minibatch of size $m$ from multiple datasets, we evenly split the minibatch by the number of training datasets $n$, meaning each dataset will contribute $\nicefrac{m}{n}$ examples to each minibatch. Accordingly, there is no notion of ``epoch'' for the unified dataset during our training, but rather only total samples seen from each dataset. For example, in a single effectual ``COCO epoch'', Mapillary will complete more than $6$ effectual epochs, as its dataset is less than  $\frac{1}{6}$th the size of COCO. For all of our quantitative studies, we train until one million crops from each dataset's images have been seen. We further train models until three million crops have been seen to share them with the community. We use a mini-batch size $m=35$. % $m=64$ for VGA and $m=128$ for QVGA. 

Image resolution is inconsistent across component datasets. For example, Mapillary contains many images of resolution $\sim$\mbox{$2000 \times 4000$}, while most ADE20K images have resolution $\sim$\mbox{$300 \times 400$}. Before training, we use 2$\times$ or 3$\times$ super-resolution \cite{Li19cvpr_FeedBackSuperRes} to first upsample the training datasets with lower resolution to a higher one (at least $1000$p).
At training time, we resize images from different datasets to a consistent resolution of $1080p$. Before training, resizing images from different datasets to a canonical size is important: during training, a crop of a fixed size is fed to the network, and such crops could represent dramatically different relative portions of the image if images' sizes differ drastically.
Specifically, in our experiments, we resize all images such that their shorter side is 1080 pixels (while preserving aspect ratios) and use a crop size of $713 \times 713$px. At test time, we resize the image to one of three different resolutions ($360/720/1080$ as the images' shorter side), perform inference, and then interpolate the prediction maps back to the original resolutions for evaluation. The resolution level ($360/720/1080$) is set per dataset. 

\subsubsection{Using the MSeg taxonomy on a held-out dataset}

At inference time, at each pixel we obtain a vector of probabilities over the unified taxonomy's $m_u$ categories. These unified taxonomy probabilities must be allocated to test dataset taxonomy buckets. For example, we have three separate probabilities in our unified taxonomy for `motorcyclist', `bicyclist', and `rider-other'. We sum these three together to compute a Cityscapes `rider' probability. We implement this remapping from $m_u$ classes to $m_t$ classes as a linear mapping $P$ from $\mathbbm{R}^{m_u}$ to $\mathbbm{R}^{m_t}$.
The weights $P_{ij}$ are binary 0/1 values and are fixed before training or evaluation; the weights are determined manually by inspecting label maps of the test datasets.  $P_{ij}$ is set to 1 if unified taxonomy class $j$ contributes to evaluation dataset class $i$, otherwise $P_{ij} =0$.

\begin{table*}[h]
    \small
    \caption{Semantic segmentation accuracy (mIoU) on MSeg training datasets. Models trained with 1 million crops. Evaluated on validation sets. \textit{Top:} performance of models trained on individual datasets. \textit{Bottom:} the same model trained on MSeg (our result). Numbers within 1\% of the best are in bold. The rightmost column is a summary measure: harmonic mean across datasets. Inference is performed at a single scale.}
    %\resizebox{\columnwidth}{!}{
      \centering
    \begin{tabular}{@{}lcccccccc@{}}
    \toprule
        Train/Test & COCO & ADE20K & Mapillary & IDD & BDD & Cityscapes & SUN & \textit{h.\ mean} \\
        \midrule
                   COCO  &  $  \mathbf{52.7} $ & $  19.1 $ & $  28.4 $ & $  31.1 $ & $  44.9 $ & $  46.9 $ & $  29.6 $ & $  32.4 $\\
                 ADE20K  &  $  14.6 $ & $  \mathbf{45.6} $ & $  24.2 $ & $  26.8 $ & $  40.7 $ & $  44.3 $ & $  36.0 $ & $  28.7 $\\
              Mapillary  &  $   7.0 $ & $   6.2 $ & $  \mathbf{53.0} $ & $  50.6 $ & $  59.3 $ & $  71.9 $ & $   0.3 $ & $   1.7 $\\
                    IDD  &  $   3.2 $ & $   3.0 $ & $  24.6 $ & $  \mathbf{64.9} $ & $  42.4 $ & $  48.0 $ & $   0.4 $ & $   2.3 $\\
                    BDD  &  $   3.8 $ & $   4.2 $ & $  23.2 $ & $  32.3 $ & $  63.4 $ & $  58.1 $ & $   0.3 $ & $   1.6 $\\
             Cityscapes  &  $   3.4 $ & $   3.1 $ & $  22.1 $ & $  30.1 $ & $  44.1 $ & $  77.5 $ & $   0.2 $ & $   1.2 $\\
               SUN RGBD  &  $   3.4 $ & $   7.0 $ & $   1.1 $ & $   1.0 $ & $   2.2 $ & $   2.6 $ & $  43.0 $ & $   2.1 $\\
 \midrule
 MSeg-w/o relabeling  &  $  50.4 $ & $  \mathbf{45.4} $ & $  \mathbf{53.1} $ & $  \mathbf{65.1} $ & $  66.5 $ & $  \mathbf{79.5} $ & $  \mathbf{49.9} $ & $  \mathbf{56.6} $\\
 \midrule
                MSeg  &  $  50.7 $ & $  \mathbf{45.7} $ & $  \mathbf{53.1} $ & $  \mathbf{65.3} $ & $  \mathbf{68.5} $ & $  \mathbf{80.4} $ & $  \mathbf{50.3} $ & $  \mathbf{57.1} $\\
 \bottomrule
    \end{tabular}%}
    \label{tab:resultsontraindatasets}
\end{table*}

\subsection{Zero-shot Quantitative Results} 
\label{sec:zeroshot_exp}
We evaluate zero-shot generalization only for semantic segmentation since our test datasets typically do not have instance labels. We use the MSeg training set to train a unified semantic segmentation model using 1 million crops. Table~\ref{tab:resultsontestdatasets} lists the results of zero-shot transfer of the model to MSeg test datasets. Note that none of these datasets were seen by the model during training. For comparison, we list the performance of corresponding models that were trained on a single training dataset, among the training datasets that were used to make up MSeg. For reference, we also list the performance of `oracle' models that were trained on the training splits of the test datasets. Note that WildDash does not have a training set, thus no `oracle' performance is provided for it. 

The results in Table~\ref{tab:resultsontestdatasets} indicate that good performance on a particular test dataset sometimes is obtained by training on a training dataset that has compatible priors. For example, training on COCO yields good performance on VOC, and training on Mapillary yields good performance on KITTI. But no individual training dataset yields good performance across all test datasets. In contrast, the model trained on MSeg performs consistently across all datasets.
This is evident in the aggregate performance, summarized by the harmonic mean across datasets. The harmonic mean mIoU achieved by the MSeg-trained model is 28\% higher than the accuracy of the best individually-trained baseline (COCO).

When compared with oracle models, training on MSeg is competitive with oracle models on many datasets, validating the importance of learning from a diverse collection of datasets. One interesting failure case is ScanNet \cite{Dai:2017CVPR:ScanNet}, as it is especially difficult in a zero-shot, cross-dataset test regime because its scenes are captured while the photographer is in motion; an oracle model trained on this data learns to account for the motion blur, yielding a 13.8 point improvement over MSeg.

\subsubsection{WildDash benchmark}
The WildDash benchmark~\cite{Zendel:2018:ECCV:WildDash} specifically evaluates the robustness of semantic segmentation models. Images mainly contain %
\begin{table}[H]
\center
\vspace{2mm}
\caption{Results from the WildDash-v1 leaderboard at the time of submission. Our model, transferred zero-shot, ranks 1st and outperforms models that utilized WildDash data during training.}
\small
    \begin{tabular}{@{}l@{\hspace{7mm}}c@{\hspace{7mm}}c@{}}
    \toprule
      &   Meta AVG mIoU  &  Seen WildDash data? \\
     \midrule
        MSeg-1080 (Ours) & $\mathbf{48.3}$ & \xmark  \\
         LDN BIN-768 \cite{Bevandic:GCPR2019:SemSegOutlierDet} & $46.9$ & \cmark  \\
         LDN OE \cite{Bevandic:GCPR2019:SemSegOutlierDet} & $42.7$ & \cmark  \\
     DN169-CAT-DUAL & $41.0$ & \cmark  \\
     AHiSS \cite{Meletis:IVS2018:heterogeneousdatasets} & $39.0$ & \xmark 
       \\
      \bottomrule
    \end{tabular}
\label{tab:wilddashresults}
\end{table}

\noindent road scenes with unusual and hazardous conditions (e.g., poor weather, noise, distortion). The benchmark is intended for testing the robustness of models trained on other datasets, and does not provide a training set of its own. 

A small set of 70 annotated images is provided for validation. The primary mode of evaluation is a leaderboard, with a testing server and a test set with hidden annotations. The main evaluation measure is Meta Average mIoU, which combines performance metrics associated with different hazards and per-frame IoU.

We submitted results from a model trained on MSeg to the WildDash test server, with multi-scale inference. Note that WildDash is not among the MSeg training sets and the submitted model has never seen WildDash images during training. The results are reported in Table~\ref{tab:wilddashresults}. Our model is ranked 1st on the leaderboard for the WildDash-v1 Benchmark. Remarkably, our model outperforms methods that were trained on multiple datasets and utilized the WildDash validation set during training. In comparison to the best prior model that (like ours) did not leverage WildDash data during training, our model improves accuracy by 9.3 percentage points: a 24\% relative improvement.

\subsection{Zero-Shot Qualitative Results}
Figure~\ref{fig:qualexamples-eachdataset} provides qualitative semantic segmentation results on images from different test datasets. Unlike the baselines, the MSeg model is successful in \textit{all} domains. 

On ScanNet, our model can accurately parse cabinets, kitchen counters, and kitchen sinks, whereas COCO cannot because COCO counter masks are not labeled accurately. Because we relabeled COCO counter masks, we are able to effectively use them for learning.  In comparison, ADE20K models are also blind to counters and Mapillary-trained models completely fail in ScanNet's indoor regime. 

%On ScanNet, our model can at times provide more accurate predictions for chairs than even the provided ground truth.

On KITTI, the ADE20K and COCO-trained models cannot recognize the sidewalk along the road edge under difficult illumination conditions, whereas the MSeg and Mapillary models can. On WildDash, our model is the only one to correctly identify trucks on both sides of a road, while the other models confuse them for buildings. On another WildDash image, our MSeg model alone can parse the road surface under snowy conditions; a Mapillary-trained model predicts the snow-covered road surface as \emph{sky}, and a COCO-trained model predicts it as \emph{wall}.

%On CamVid, the Mapillary and COCO-trained models incorrectly predict sidewalk on the road surface; ADE20K and COCO-trained models have no notion of rider and mistake bicyclists for pedestrians. On Pascal VOC, our model is the only one to correctly identify a person standing on an airplane's mobile staircase; an ADE20K-trained model predicts a boat, and a Mapillary model sees a car. On another Pascal image, ADE20K has no horse class, and the corresponding model cannot identify it.

\subsection{Performance on Training Datasets}
\label{sec:training_performance}
Table \ref{tab:resultsontraindatasets} lists the accuracy of trained models on the MSeg training datasets. We test on the validation sets and compute IoU on a subset of classes that are jointly present in the dataset and MSeg's taxonomy. Except for Cityscapes and BDD100K, results on validation sets of all training datasets are not directly comparable to the literature since the MSeg taxonomy involves merging multiple classes. As expected, individually-trained models generally demonstrate good accuracy when tested on the same dataset: a model trained on COCO performs well on COCO, etc. Remarkably, the MSeg model matches or outperforms the respective individually-trained models across almost all datasets. On a number of training datasets, the MSeg-trained model attains higher accuracy than the individually-trained model, sometimes by a significant margin.
The aggregate performance of the MSeg model is summarized by the harmonic mean across datasets. It is 76\% higher than the best individually-trained baseline (COCO).

\subsection{Ablation Study for Relabeling}
Table~\ref{tab:naive-unified-relabeling-comparison} reports a controlled evaluation of two of our contributions: the unified taxonomy (Section~\ref{sec:unified-tax-principles}) and compatible relabeling (Section~\ref{sec:amt}). The `Naive merge' baseline is a model trained on a composite dataset that uses a naively merged taxonomy in which the classes are a union of all training classes, and each test class is only mapped to an universal class if they share the same name. 
%We train the models using this naive taxonomy and call it \emph{Naive Merge}.
The `MSeg (w/o relabeling)' baseline uses the unified MSeg taxonomy, but does not use the manually-relabeled data for split classes (Section~\ref{sec:amt}). The model trained on the presented composite dataset (`MSeg') achieves better performance than the baselines. Using MSeg's unified taxonomy gives a large boost over the naive taxonomy, and relabeling the data further helps.
%In Table~\ref{tab:naive-unified-relabeling-comparison}, we also note that relabeling appears to slightly hurt performance on VOC (69.2\% vs. 69.4\% w/o relabeling). However, this 0.2\% mIoU drop may be attributed to noise; we routinely saw such slight variation as we repeated experiments throughout our work. 

\begin{table}[H]
    \centering
    \small
    \caption{Controlled evaluation of unified taxonomy and mask relabeling. Zero-shot transfer to MSeg test datasets. Both contributions make a positive impact on generalization accuracy.}
    \begin{adjustbox}{max width=\columnwidth}
            \begin{tabular}{@{}l@{\hspace{1mm}}c@{\hspace{2mm}}c@{\hspace{2mm}}c@{\hspace{2mm}}c@{\hspace{2mm}}c@{\hspace{2mm}}c@{\hspace{2mm}}c@{}}
    \toprule
       Train/Test & VOC & Context & CamVid & WildDash   & KITTI  & ScanNet & \textit{h.\ mean} \\
    \midrule
        %Naive merge  &  $  51.9 $ & $  23.8 $ & $  56.2 $ & $  59.7 $ & $  62.6 $ & $  43.4 $ & $  44.5 $\\
        Naive merge   &  $  17.8 $ & $  19.4 $ & $  56.3 $ & $  55.9 $ & $  61.6 $ & $  45.6 $ & $  33.1 $\\
        MSeg w/o relabeling & $\mathbf{70.2}$ & $\mathbf{42.7}$ & $82.0$ & $\mathbf{62.7}$ & $65.5$ & $43.2$ & $57.6$ \\
        MSeg  & $\mathbf{70.7}$ & $\mathbf{42.7}$ & $\mathbf{83.3}$ & $\mathbf{62.0}$ & $\mathbf{67.0}$ & $\mathbf{48.2}$ & $\mathbf{59.2}$ \\
\bottomrule
    \end{tabular}
    \end{adjustbox}
    \label{tab:naive-unified-relabeling-comparison}
\end{table}

\subsection{Performance in the 2020 Robust Vision Challenge}

\begin{figure*}[t]
\centering
\begin{tabular}{@{}c@{}}
\includegraphics[width=0.95\linewidth]{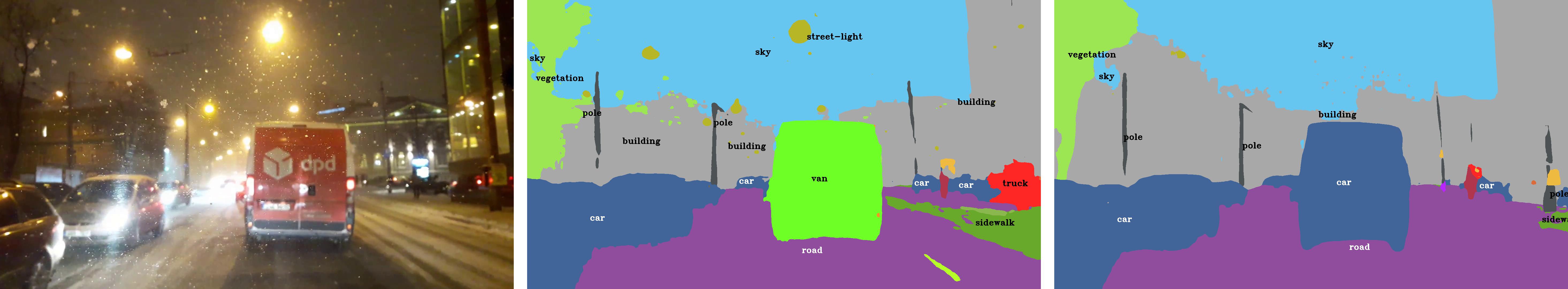}\\
\includegraphics[width=0.95\linewidth]{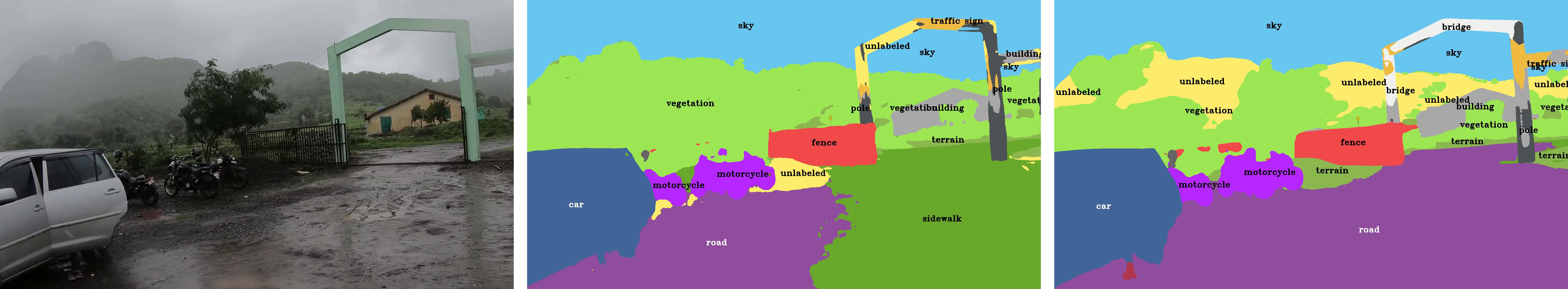}\\
\includegraphics[width=0.95\linewidth]{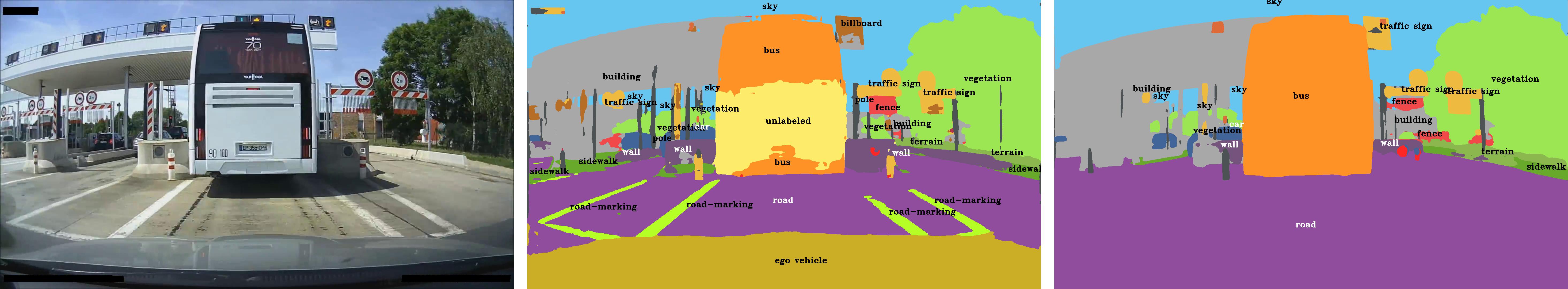}\\
\includegraphics[width=0.95\linewidth]{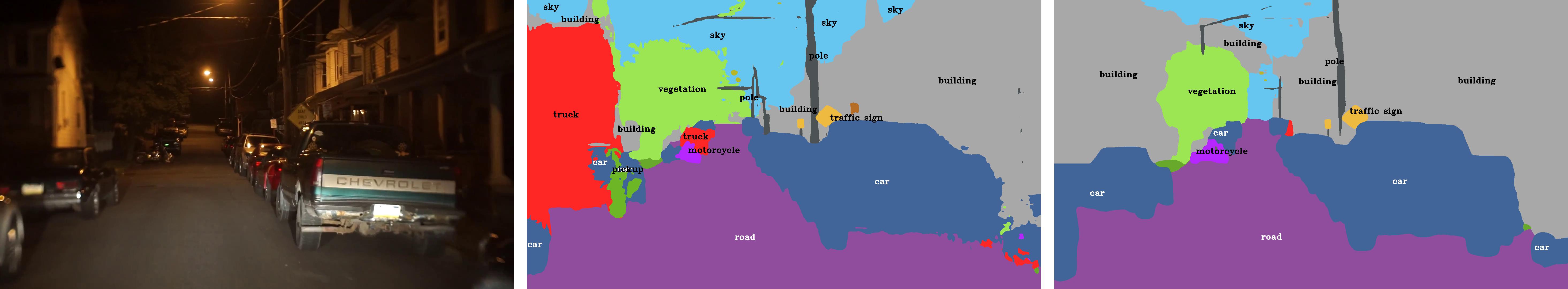}\\
\end{tabular}
  \begin{tabular}{@{}p{0.30\linewidth}@{\hspace{7mm}}p{0.30  \linewidth}@{\hspace{7mm}}p{0.30\linewidth}@{}}
        \multicolumn{1}{@{}c@{}}{Input image} &
          \multicolumn{1}{@{}c@{}}{SwiftNet model} &  \multicolumn{1}{@{}c@{}}{MSeg model}\\ && \\
 \end{tabular}
 \vspace{-5mm}
 \caption{Qualitative results from SwiftNet (RVC 2020 champion) and our MSeg model (RVC 2020 runner-up without any retraining on RVC datasets) on images from the \textbf{WildDash-v2} RVC test dataset. Taxonomies greatly influence predictions; SwiftNet and WildDash-v2 introduce a separate `van' class (see row 1), whereas we group `van' instances with the `car' category. SwiftNet and WildDash-v2 also introduce a separate `lane marking' class (see row 3), which we merge into `road'. Note that SwiftNet's taxonomy incorporates an `ego-vehicle' class, whereas we treat `ego-vehicle' as `unlabeled' when training MSeg models (See row 3).}
 \label{fig:qualswiftexamples-eachdataset}
 \vspace{2mm}
 \captionof{table}{Robust Vision Challenge (RVC) 2020 results in the semantic segmentation track, as measured by class mIoU. We report zero-shot cross-dataset generalization in \red{red} and testing on the test split of datasets that were already seen at training time in \blue{blue}. The best number in each column is highlighted in bold.} %Competition participants did not enter results for all evaluation datasets (marked via $-$).}
\begin{adjustbox}{max width=\textwidth}
\begin{tabular}{@{}lcccccccc@{}}
\toprule
  & \multicolumn{7}{c}{ \textsc{RVC Test Datasets} } \\
\textsc{Method Name}    &  \textsc{ADE20K} & \textsc{Cityscapes} &  \textsc{KITTI}  & \textsc{Mapillary} & \textsc{ScanNet} & \textsc{VIPER} & \textsc{WildDash-v2}  \\
\midrule
\textsc{SwiftNet (SN\_RN152pyrx8\_RVC)} \cite{Orsic20arxiv_MultiDomainRVC20} & \blue{31.12}   & \blue{74.7}      & \blue{\textbf{63.89}}     & \blue{\textbf{40.43}}     & \blue{\textbf{54.6}}      & \blue{\textbf{62.5}}  & \blue{\textbf{42.29}}  \\
\textsc{MSeg1080\_RVC}   & \blue{\textbf{33.18}}   & \blue{\textbf{80.7}}   & \red{62.64}    & \blue{34.19}   & \red{48.5}   & \red{40.7}  & \red{34.71} \\
\bottomrule
\end{tabular}
\label{tab:rvcresults}
\end{adjustbox}
\end{figure*}

\begin{table*}[t]
\centering
\caption{Semantic segmentation accuracy (mIoU) on MSeg train and test datasets after a longer training. Performance of models trained on MSeg using 3 million crops. Numbers within 1\% of the best are in bold. The rightmost column is a summary measure: harmonic mean across datasets. Inference is performed at a single scale.}
\vspace{-2mm}
    \label{tab:traintest3m}
\small
\resizebox{\textwidth}{!}{
\begin{tabular}{@{}l@{\hspace{2mm}}c@{\hspace{2mm}}c@{\hspace{2mm}}c@{\hspace{2mm}}c@{\hspace{2mm}}c@{\hspace{2mm}}c@{\hspace{2mm}}c@{\hspace{5mm}}c@{\hspace{2mm}}c@{\hspace{2mm}}c@{\hspace{2mm}}c@{\hspace{2mm}}c@{\hspace{2mm}}c@{\hspace{2mm}}c@{\hspace{2mm}}c@{}}
\toprule
    & \multicolumn{7}{c}{Test Datasets (Zero-Shot Generalization)} & \multicolumn{8}{c}{Train Datasets} \\
     \cmidrule(lr\hspace{5mm}){2-8} \cmidrule(r){9-16}
Train/Test & VOC & Context & CamVid & WildDash   & KITTI  & ScanNet & \textit{h.\ mean} & COCO & ADE20K & Mapillary & IDD & BDD & Cityscapes & SUN & \textit{h.\ mean} \\ \midrule
      MSeg-3m-480p  &  $  \mathbf{76.4} $ & $  \mathbf{45.9} $ & $  81.2 $ & $ \mathbf{62.7} $ & $  \mathbf{68.2} $ & $  \mathbf{49.5} $ & $  \mathbf{61.2} $  &  $  \mathbf{56.1} $ & $  \mathbf{49.6} $ & $  53.5 $ & $  64.5 $ & $  67.8 $ & $  79.9 $ & $  49.2 $ & $  58.5 $ \\
      MSeg-3m-720p  &  $  74.7 $ & $  44.0 $ & $  83.5 $ & $  60.4 $ & $  67.9 $ & $  47.7 $ & $  59.8 $ &  $  53.3 $ & $  48.2 $ & $  53.5 $ & $  64.8 $ & $  68.6 $ & $  79.8 $ & $  49.3 $ & $  57.8 $\\
     MSeg-3m-1080p  &  $  72.0 $ & $  44.0 $ & $  \mathbf{84.5} $ & $  59.9 $ & $  66.5 $ & $  \mathbf{49.5} $ & $  59.8 $ &  $  53.6 $ & $  49.2 $ & $  \mathbf{54.9} $ & $  \mathbf{66.3} $ & $  \mathbf{69.1} $ & $  \mathbf{81.5} $ & $  \mathbf{50.1} $ & $  \mathbf{58.8} $\\ \bottomrule
\end{tabular}}
\end{table*}

As an extreme generalization experiment, we submit an MSeg model to the 2020 Robust Vision Challenge (RVC) semantic segmentation competition. It is important to clarify that RVC uses a different set of evaluation classes than our model was trained for. Moreover, RVC allows training on seven datasets, four of which are not among the MSeg training datasets. Hence, our model uses a different taxonomy and does not utilize the majority of the datasets in the RVC, making this evaluation significantly biased against our submission. Success in such a setting is only possible if our model can truly generalize beyond MSeg. Surprisingly, our submission ranks 2nd without any retraining. In the rest of this section, we analyze this experiment in more detail.

The RVC is a biennial workshop and set of competitions held in conjunction with the ECCV conference. Prior to the start of the competition, we provided our MSeg codebase and models to the organizers of the competition, who used it to prepare the RVC devkit for competitors. A new dataset, WildDash-v2, was introduced for the August 2020 competition, replacing WildDash-v1 and introducing new evaluation classes such as `pickup-truck', `lane marking', and `van'.  Rather than measuring robustness via cross-dataset zero-shot generalization, the RVC competition uses the test split of 7 training datasets, as opposed to the validation split we use in Section~\ref{sec:training_performance}. Therefore, submissions trained in an ``oracle'' mode (see Section~\ref{sec:zeroshot_exp}) are permitted, although

\begin{table}[H]
\vspace{1mm}
\caption{Quantifying the effect of taxonomy resolution on performance on Mapillary Vistas in the RVC 2020 challenge. A total of 19 Mapillary Vistas (Public) classes are merged in our universal taxonomy. We report mIoU of competing methods for our matching taxonomy (\textsc{Mapillary-46}) and the finer-grained taxonomy (\textsc{Mapillary-65}).}
\begin{adjustbox}{max width=\columnwidth}
\begin{tabular}{@{}lcc@{}}
\toprule
\textsc{Method Name}    &  \textsc{Mapillary-65} &  \textsc{Mapillary-46} \\
\midrule
\textsc{SwiftNet (SN\_RN152pyrx8\_RVC)} \cite{Orsic20arxiv_MultiDomainRVC20} & \textbf{40.43} & 43.53 \\
\textsc{MSeg1080\_RVC}  & 34.19 & \textbf{48.31} \\
\bottomrule
\end{tabular}
\end{adjustbox}
%%%
\label{tab:rvctaxonomycomparison}
\end{table}
\noindent  each submission must be a single, unified model. The RVC creates an aggregate rank for methods based on their rank per dataset on 7 pre-defined evaluation datasets \cite{Schulze11scw_NewElectionMethod}.

The competition winner, SwiftNet~\cite{Orsic20arxiv_MultiDomainRVC20}, was specifically trained on the 7 RVC test datasets (see Table \ref{tab:rvcresults}); as a result, SwiftNet suffered from no domain gap, whereas our MSeg model saw only 3 of the 7 datasets during training. For models that are trained and evaluated on the same datasets, we find MSeg outperforms SwiftNet in all cases except Mapillary Vistas as seen in Table \ref{tab:rvcresults}.
For Mapillary, relative resolution of SwiftNet's taxonomy and our taxonomy are different. We merged 19 fine-grained Mapillary classes into our universal classes, causing us to receive a score of zero when formulating a mapping backwards. In order to quantify the impact of this taxonomy difference, we re-compute the mIoU by ignoring these 19 classes in Table \ref{tab:rvctaxonomycomparison}. As seen from the table, when a matching taxonomy is used,  our mIoU score is 5\% higher than SwiftNet's.  We visualize this taxonomy difference qualitatively in Figure~\ref{fig:qualswiftexamples-eachdataset} for the WildDash-v2.

\begin{figure*}[!h]
\centering
\begin{tabular}{@{}c@{\hspace{2mm}}c@{}}
\begin{tabular}{@{}c@{}}
\includegraphics[width=0.49\linewidth]{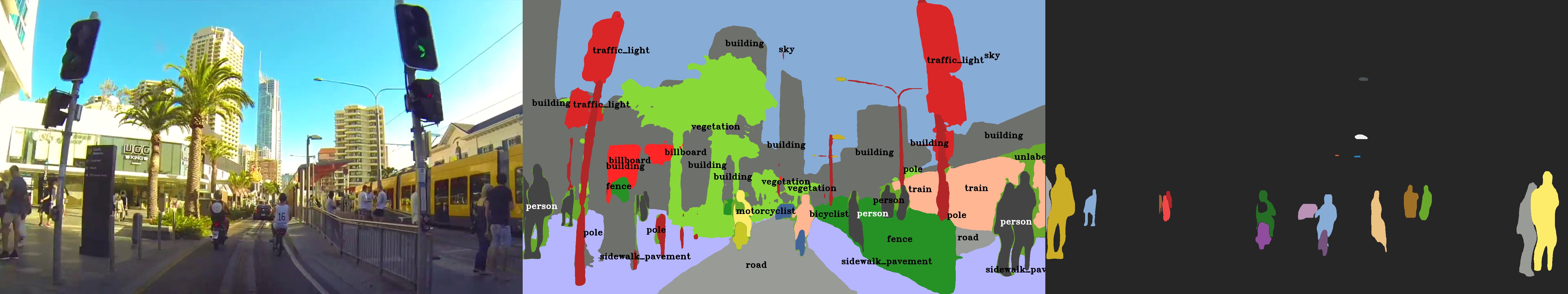} \\
\includegraphics[width=0.49\linewidth]{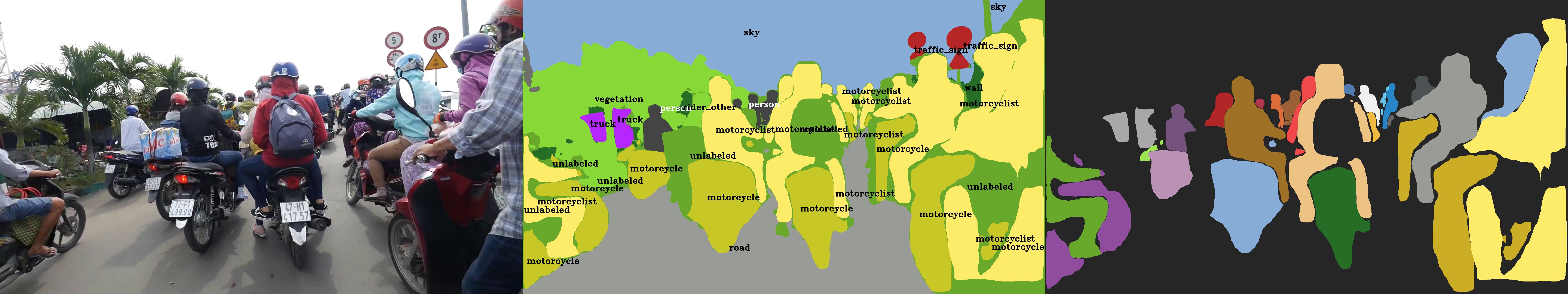}  \\
\includegraphics[width=0.49\linewidth]{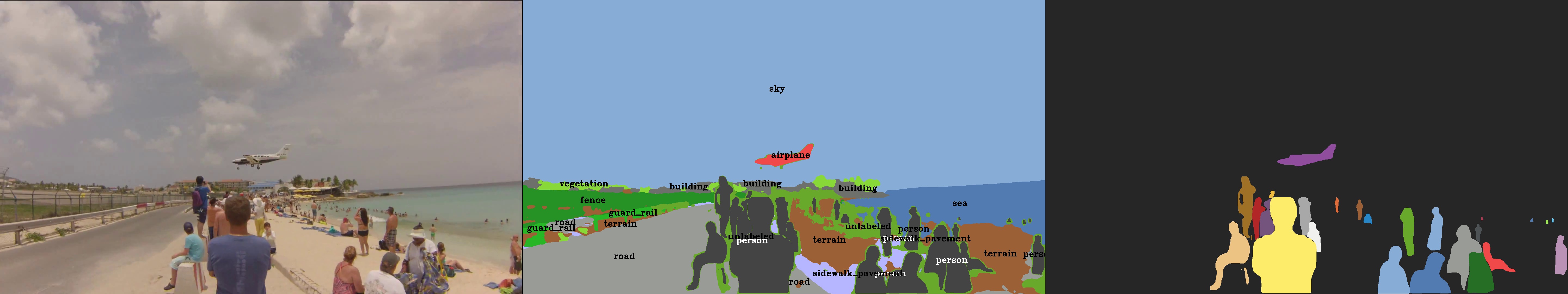} \\
\includegraphics[width=0.49\linewidth]{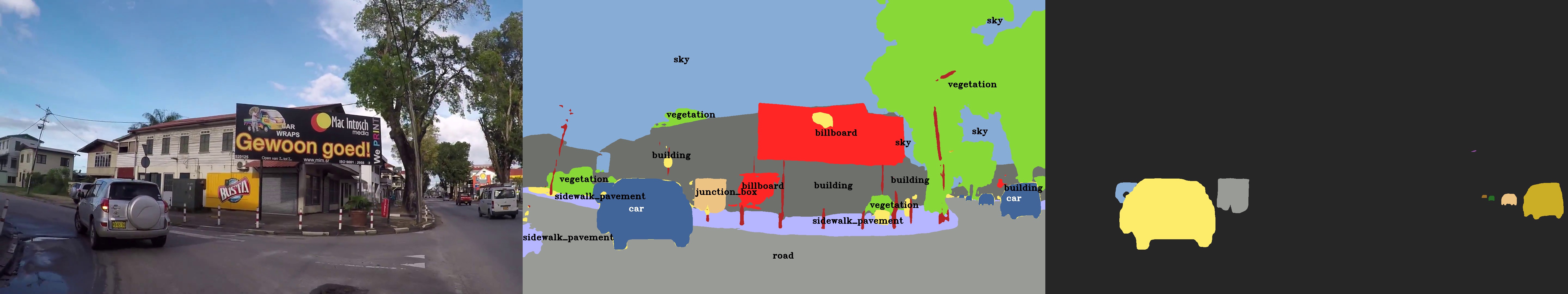}\\
  \begin{tabular}{@{}p{0.185\linewidth}@{\hspace{1mm}}p{0.08 \linewidth}@{\hspace{2mm}}p{0.185\linewidth}@{}}
        \multicolumn{1}{@{}c@{}}{Input image} &
          \multicolumn{1}{@{}c@{}}{Semantic Output} &  \multicolumn{1}{@{}c@{}}{Instance Output}\\ && \\
 \end{tabular}
\end{tabular} &
\begin{tabular}{@{}c@{}}
\includegraphics[width=0.49\linewidth]{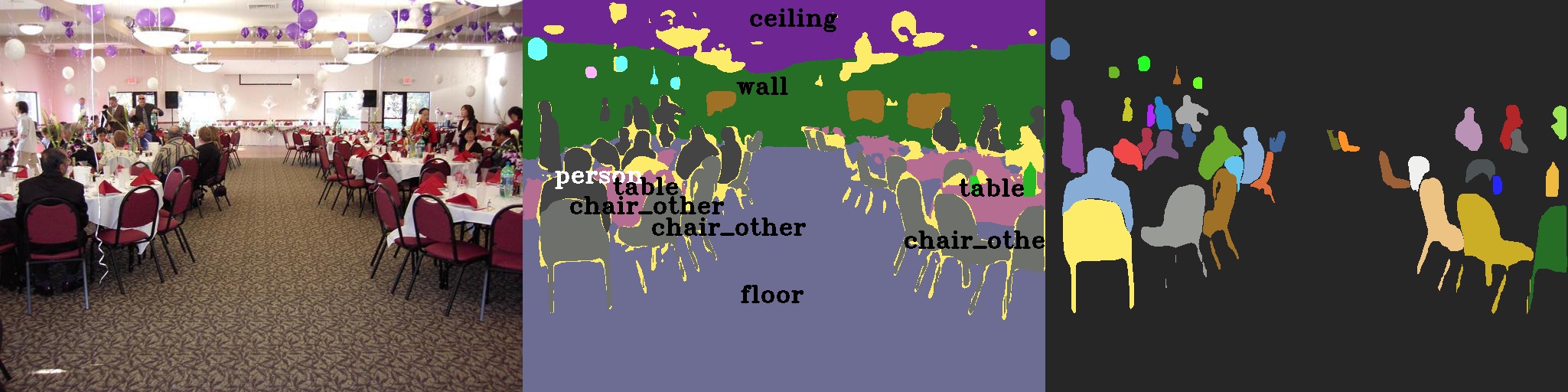}
\\ \includegraphics[width=0.49\linewidth]{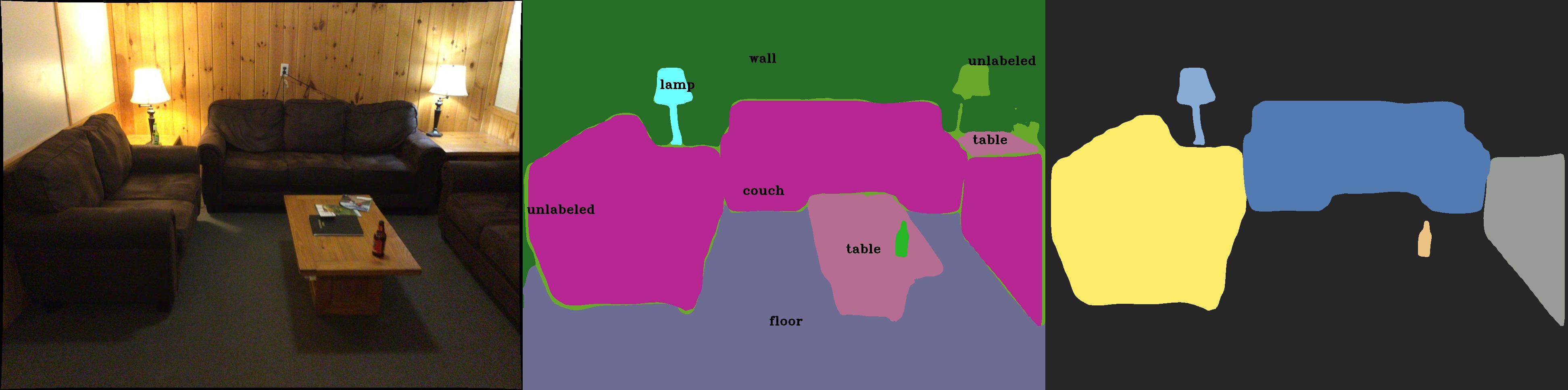} 
 \\
\includegraphics[width=0.49\linewidth]{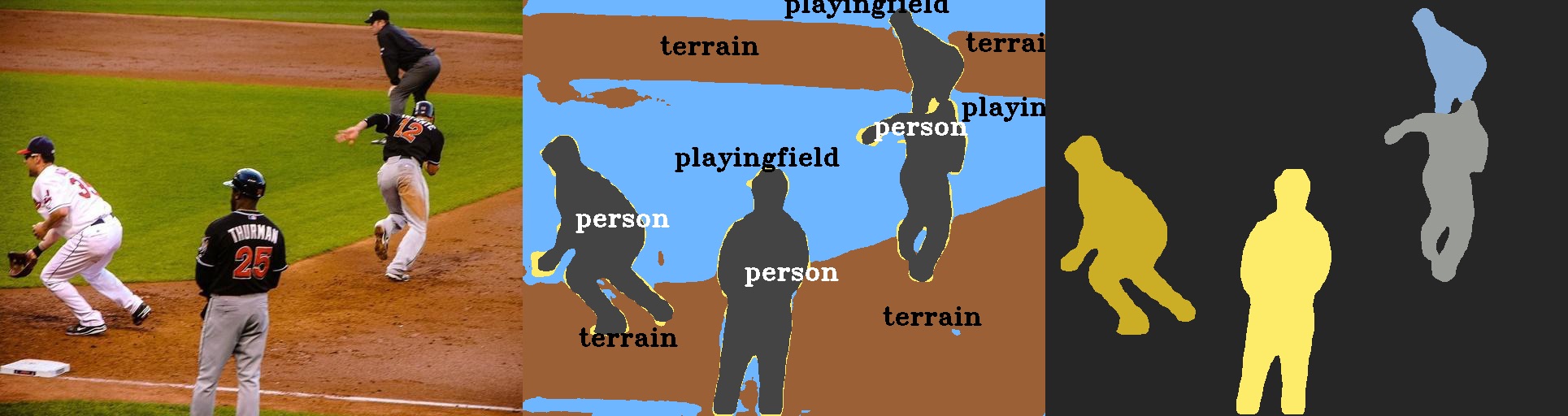} \\
  \begin{tabular}{@{}p{0.185\linewidth}@{\hspace{1mm}}p{0.08 \linewidth}@{\hspace{2mm}}p{0.185\linewidth}@{}}
        \multicolumn{1}{@{}c@{}}{Input image} &
          \multicolumn{1}{@{}c@{}}{Semantic Output} &  \multicolumn{1}{@{}c@{}}{Instance Output}\\ && \\
 \end{tabular}
\end{tabular}
\end{tabular}
 \vspace{-5mm}
 \caption{Qualitative results of our MSeg-720 panoptic model. Images sampled from the test splits of WildDash v2, ADE20K, and ScanNet.}
 \label{fig:panopticresults}
%\end{figure*}
%\begin{table*}[b]
\vspace{5mm}
    \captionof{table}{Instance segmentation accuracy on MSeg training datasets. (Evaluated on validation sets.)  Performance of models trained on MSeg. Inference is performed at a single scale.}
    \small
    \begin{tabular}{@{}lccccc@{\hspace{7mm}}ccccc@{}}
    \toprule
    & \multicolumn{5}{c}{Box Average Precision} & \multicolumn{5}{c}{Mask Average Precision} \\
    \cmidrule(lr\hspace{7mm}){2-6} \cmidrule(r){7-11}
        Resolution &   COCO & ADE20K & Mapillary & IDD  & Cityscapes & COCO & ADE20K & Mapillary & IDD  & Cityscapes\\
        \midrule
        MSeg-3m-480p & $35.1$ & $\mathbf{26.2}$ & $13.9$ & $33.4$ & $\mathbf{30.2}$ & $32.1$ & $\mathbf{21.7}$ & $11.9$ & $29.4$ & $24.6$ \\
        MSeg-3m-720p & $\mathbf{35.8}$ & $\mathbf{26.2}$ & $15.7$ & $\mathbf{35.4}$ & $\mathbf{30.2}$ & $\mathbf{32.6}$ & $21.3$ & $13.7$ & $\mathbf{31.2}$ & $\mathbf{24.8}$ \\
        MSeg-3m-1080p & $35.2$ & $24.7$ & $\mathbf{17.0}$ & $\mathbf{35.4}$ & $\mathbf{30.2}$ & $32.0$ & $20.3$ & $\mathbf{15.1}$ & $\mathbf{31.2}$ & $\mathbf{24.8}$ \\
 \bottomrule
    \end{tabular}
    \label{tab:instance}
\end{figure*}

\subsection{Practical Semantic Segmentation for Practitioners}
Our main motivation behind MSeg is developing and sharing practical multi-domain models which practitioners can use without any additional training. In order to ease this adoption, we perform two crucial steps. Firstly, we go beyond the training duration of 1 million crops and obtain better models with longer training. Specifically, we train models until 3 million crops per dataset are seen. Secondly, we go beyond 1080p resolution, and train 480p and 720p resolution models. This is especially crucial for mobile applications where computation is scarcer. In order to understand the accuracy-computation trade-off, we analyze the accuracy and computational efficiency of these models. We tabulate these results in Table~\ref{tab:traintest3m}. 

One interesting observation is the superior zero-shot generalization of the 480p model in the test datasets. On the other hand, the higher resolution is helpful for performance in training datasets. Although this is promising for practitioners as they could use the low resolution models and obtain the best of both worlds, it is somewhat puzzling. Although explaining this discrepancy is vastly beyond the scope of this paper, we state a conjecture which might help explain this phenomenon.
% we state two conjectures which might help explain this phenomenon. Touvron et al.~\cite{Touvron_2019} demonstrated that data augmentation creates a mismatch between train and test distributions favoring lower-resolution training. Although this perspective is promising, it does not explain the discrepancy between zero-shot and within-distribution generalization. Another 
One promising perspective would be the distributional distances, as they are strongly related with the out-of-distribution generalization \cite{Shai2010}. It is reasonable to conjecture that the distribution of different training sets gets closer in a lower-resolution setting. As a supporting thought experiment, consider classifying an image into respective datasets (similar to \cite{Torralba:2011:DatasetBias}). It would be harder to classify which dataset the image is coming from at a lower resolution.

In order to quantify the computational efficiency,
we provide an analysis of the feedforward inference runtime of our semantic segmentation models in Table \ref{tab:runtime-analysis}. 

\begin{wraptable}{r}{35mm}
\vspace{-4mm}
\caption{Average frame rate for different resolutions.}
\centering
\begin{tabular}{@{\hspace{2mm}}l@{\hspace{5mm}}c@{\hspace{1mm}}c@{\hspace{2mm}}}
\toprule         
& RTX 2080 Ti  \\
\midrule 
420p  & 27.19 \\
720p  & 19.29 \\
1080p & 15.64\\
\bottomrule
\end{tabular}
\label{tab:runtime-analysis}
\vspace{-5mm}
\end{wraptable}
We average runtime over 100 single-scale forward passes. Real-time single-scale inference is possible on an NVIDIA RTX 2080 Ti desktop GPU. We only perform single-scale analysis as these experiments are performed on specially optimized single-scale inference code which we share with the community\footnote{All trained models as well as training and inference codes are available at \textcolor{blue}{\url{http://github.com/mseg-dataset/mseg-semantic}}}.

\subsection{Going Beyond Semantic Segmentation}
Semantic segmentation problem has no sense of an object or instance. Instance and panoptic segmentation extend this problem to introduce the concept of objects. Instance segmentation segments an image into semantically meaningful instances. However, it only considers object categories, not ``stuff'. On the other hand, panoptic segmentation jointly performs instance segmentation and semantic segmentation. 

We train instance and panoptic segmentation models using MSeg since the vast majority of training datasets have instance labels. However, we only evaluate these models on the MSeg training datasets since our test datasets typically do not have instance labels. During our training and evaluation, we exclude BDD and SUN RGBD because they do not provide instance labels. We use validation sets of these datasets in our evaluation. To differentiate between ``thing'' and ``stuff'' classes in the universal taxonomy, we deem the class to be a ``thing'' if all classes mapped to that universal class are ``thing'' classes in their original datasets. This step is necessary since instance segmentation detects and segments ``thing'' objects, while panoptic segmentation will also segment ``stuff'' pixels. 

\begin{table}[h]
    \caption{Panoptic segmentation accuracy (PQ) on MSeg training datasets. (Evaluated on validation sets.) }
    \vspace{-2mm}
    \small
    \centering
    \begin{tabular}{@{}l@{\hspace{2mm}}c@{\hspace{2mm}}c@{\hspace{2mm}}c@{\hspace{2mm}}c@{\hspace{2mm}}c@{}}
    \toprule
        Resolution & COCO & ADE20K & Mapillary & IDD  & Cityscapes \\
        \midrule
    MSeg-3m-480p  &  $35.7$ & $32.7$ & $14.0$ & $\mathbf{47.5}$ & $49.5$ \\
    MSeg-3m-720p  &  $35.7$ & $32.9$ & $11.9$ & $44.2$ & $50.8$ \\
   MSeg-3m-1080p  &  $\mathbf{38.8}$ & $\mathbf{33.7}$ & $\mathbf{14.9}$ & $45.0$ & $\mathbf{51.1}$ \\
 \bottomrule
    \end{tabular}
    \label{tab:panoptic}
\end{table}

We evaluate instance segmentation performance by reporting box average precision and mask average precision in Table~\ref{tab:instance} for various resolutions. Although the 720p resolution is clearly superior to 480p, 1080p does not improve upon 720p. We hypothesize this surprising result can be attributed to the mismatch between train and test distributions due to data augmentation following Touvron \emph{et al.}~\cite{Touvron_2019}.

We further evaluate the resulting models in the joint instance and semantic segmentation task called panoptic segmentation. We use panoptic quality as a metric and tabulate the results in Table~\ref{tab:panoptic}. We also present qualitative results in Figure~\ref{fig:panopticresults}. Although all resolutions perform similarly in our quantitative analysis, high resolution models perform best. Moreover, the qualitative results in Figure~\ref{fig:panopticresults} are promising.

Although answering the question of whether MSeg-relabeling improves instance and panoptic segmentation is not directly possible due to the lack of instance labels in many datasets we use, the results presented in this section suggests that practitioners can use these models in domain-free sense.

\section{Conclusion}

We presented a composite dataset for multi-domain scene understanding. To construct the composite dataset, we reconciled the taxonomies of seven semantic segmentation datasets. In cases where categories needed to be split, we performed large-scale mask relabeling via the Mechanical Turk platform. We showed that the resulting composite dataset enables training a unified semantic segmentation model that delivers consistently high performance across domains. The trained model generalizes to previously unseen datasets and was ranked first on the WildDash leaderboard for robust semantic segmentation, with no exposure to WildDash data during training. Moreover, we submitted the resulting models to the Robust Vision Challange (RVC) without any additional training as an extreme generalization experiment. Our model performed competitively and ranked second. MSeg training sets only include three out of the seven datasets in the RVC, making this competitive performance surprising and promising. In order to guide practical deployment of our models, we also evaluated the role of resolution in the trade-off between computational efficiency and accuracy. Finally, we trained instance and panoptic segmentation models to go beyond semantic segmentation. We see the presented work as a step towards broader deployment of robust computer vision systems and hope that it will support future work on zero-shot generalization. Code, data, and trained models are available at \textcolor{blue}{\url{https://github.com/mseg-dataset}}.

\textbf{Acknowledgments.} We thank Marin Oršić, Petra Bevandić, and Siniša Šegvić for sharing their SwiftNet label map predictions from the 2020 Robust Vision Challenge.

\clearpage

\ifCLASSOPTIONcompsoc
 % \section*{Acknowledgments}
\else
  %\section*{Acknowledgment}
\fi

\ifCLASSOPTIONcaptionsoff
  \newpage
\fi



\bibliographystyle{IEEEtran}
\bibliography{IEEEabrv,egbib.bib}

% {\small
% \bibliographystyle{ieee_fullname}
% \bibliography{egbib}


% \end{thebibliography}


\begin{IEEEbiography}[{\includegraphics[width=1in,height=1.5in,clip,keepaspectratio]{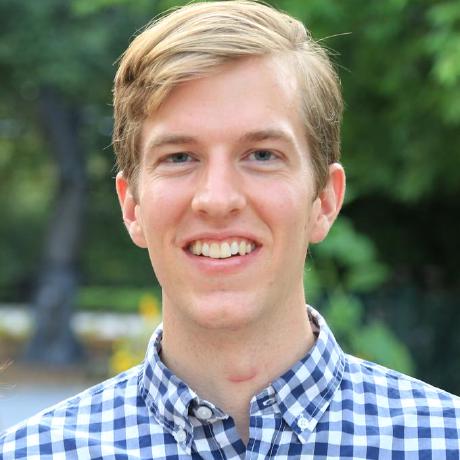}}]{John Lambert}
is a Ph.D student in computer vision and robotics at the Georgia Institute of Technology, advised by Dr. James Hays and Dr. Frank Dellaert. He received his M.S. and B.S. degrees in Computer Science in 2018 and 2017 from Stanford University, where he was advised by Dr. Silvio Savarese. His research interests center around scene understanding and 3d reconstruction and mapping for mobile robotics.
\end{IEEEbiography}

\begin{IEEEbiography}[{\includegraphics[width=1in,height=1.5in,clip,keepaspectratio]{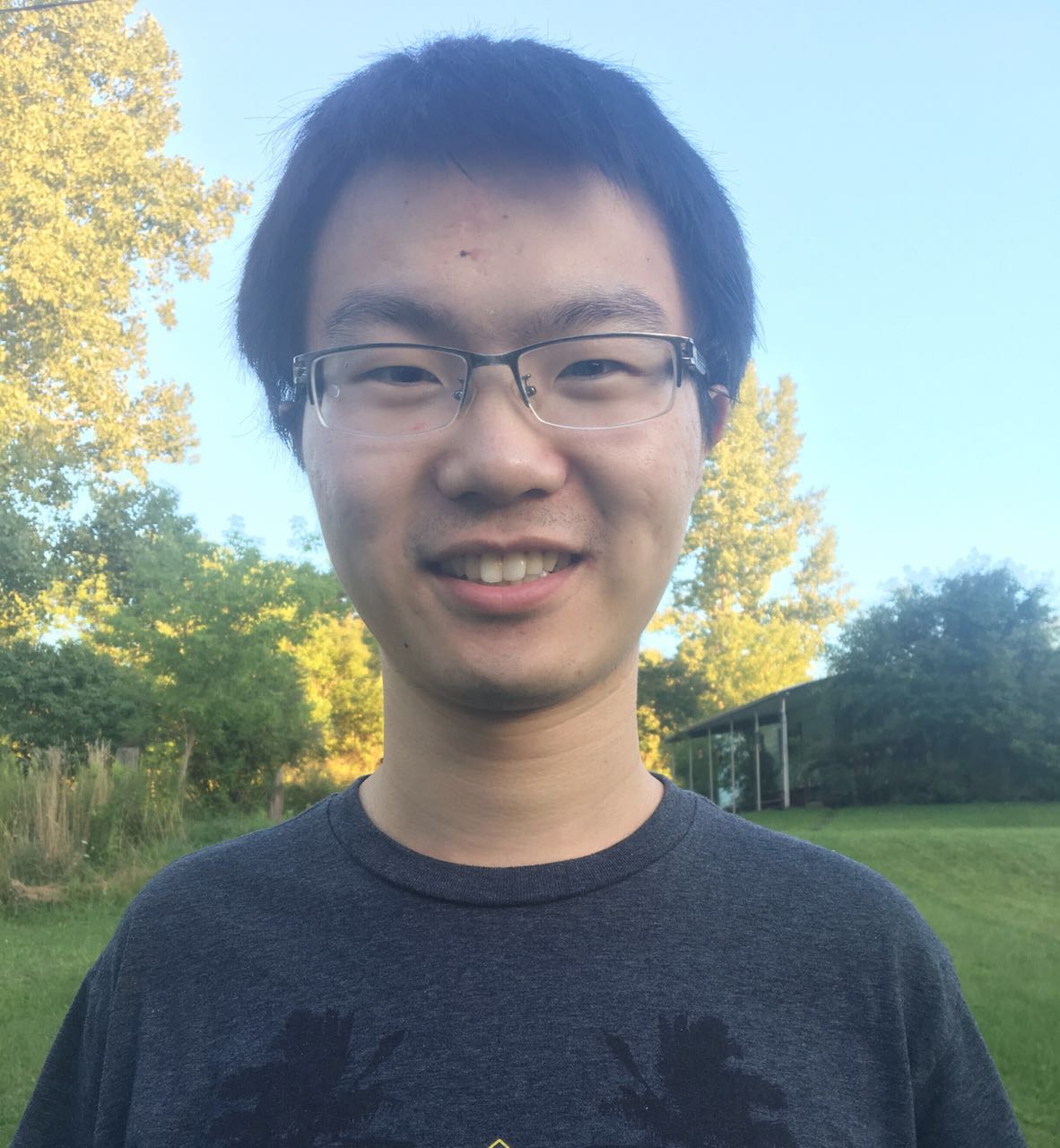}}]
{Zhuang Liu} is a Ph.D. student at University of California, Berkeley, advised by Prof. Trevor Darrell. He received his B.S. degree in Computer Science from Institute for Interdisciplinary Information Sciences (Yao Class), Tsinghua University in 2017. He spent time as an intern at Cornell University, Intel Labs and Adobe Research. His research mainly focuses on the computational and data efficiency of deep learning.
\end{IEEEbiography}

\begin{IEEEbiography}[{\includegraphics[width=1in,height=1.5in,clip,keepaspectratio]{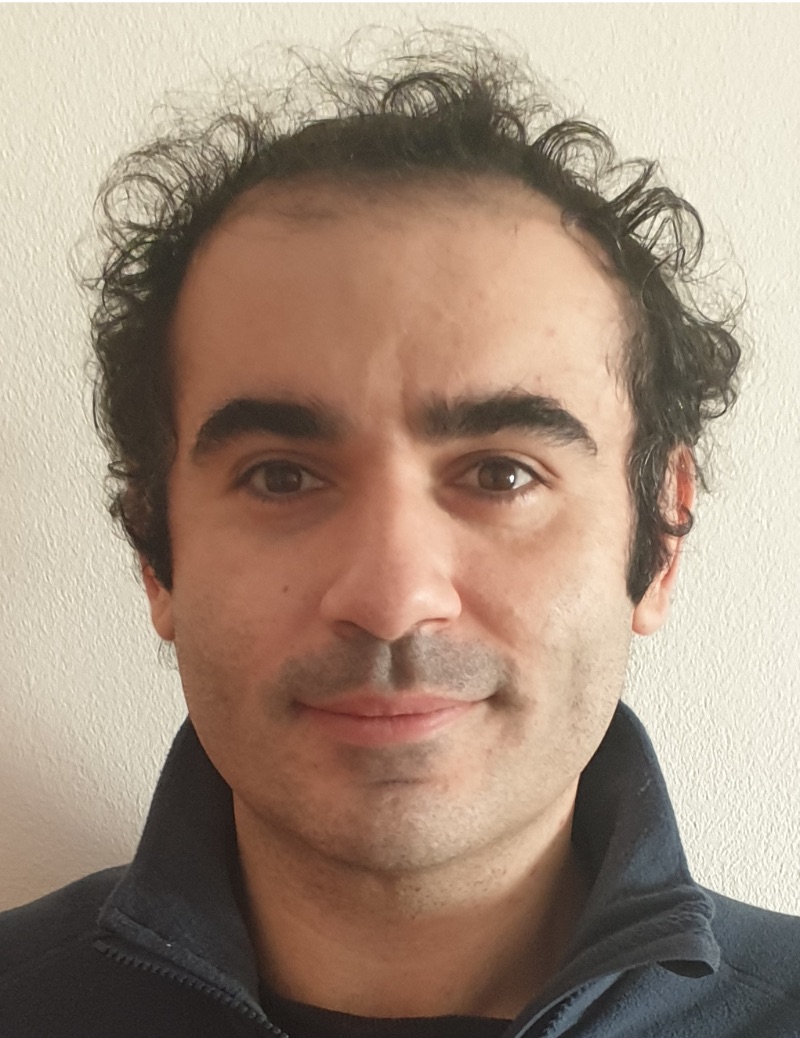}}]{Ozan Sener}
is a research scientist at Intel Labs. In 2017, he earned his Ph.D. degree from Cornell University. Between 2016 and 2018, he worked as a visiting Ph.D. student and later as a PostDoc at Stanford University. He is interested in developing machine learning methods and theory, with applications in computer vision and robotics. His past research focused on transfer and multi-task learning, derivative-free optimization, and unsupervised learning. \end{IEEEbiography}

\begin{IEEEbiography}[{\includegraphics[width=1.0in,clip,keepaspectratio]{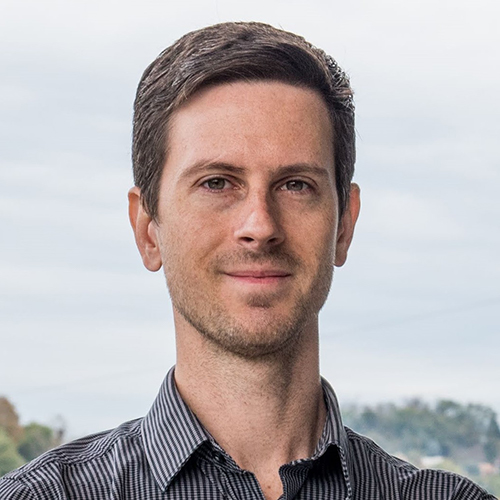}}]{James Hays}
is an associate professor of computing at Georgia Institute of Technology and a Principal Scientist at Argo AI. James received his Ph.D. from Carnegie Mellon University in 2009 and was a postdoc at Massachusetts Institute of Technology. James's research interests span computer vision, robotics, and machine learning. James is the recipient of the NSF CAREER award and Sloan Fellowship.
\end{IEEEbiography}

\begin{IEEEbiography}[{\includegraphics[width=1.0in,clip,keepaspectratio]{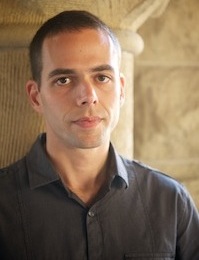}}]{Vladlen Koltun}
is the Chief Scientist for Intelligent Systems at Intel. He directs the Intelligent Systems Lab, which conducts high-impact basic research in computer vision, machine learning, robotics, and related areas. He has mentored more than 50 PhD students, postdocs, research scientists, and PhD student interns, many of whom are now successful research leaders.
\end{IEEEbiography}

\newpage
\section*{\Large{Appendix}}

In this appendix, we provide additional details and experimental results.

\section{Relabeling Interface and Workflow}
\begin{figure}[ht]
    \centering
    \includegraphics[width=0.9\columnwidth]{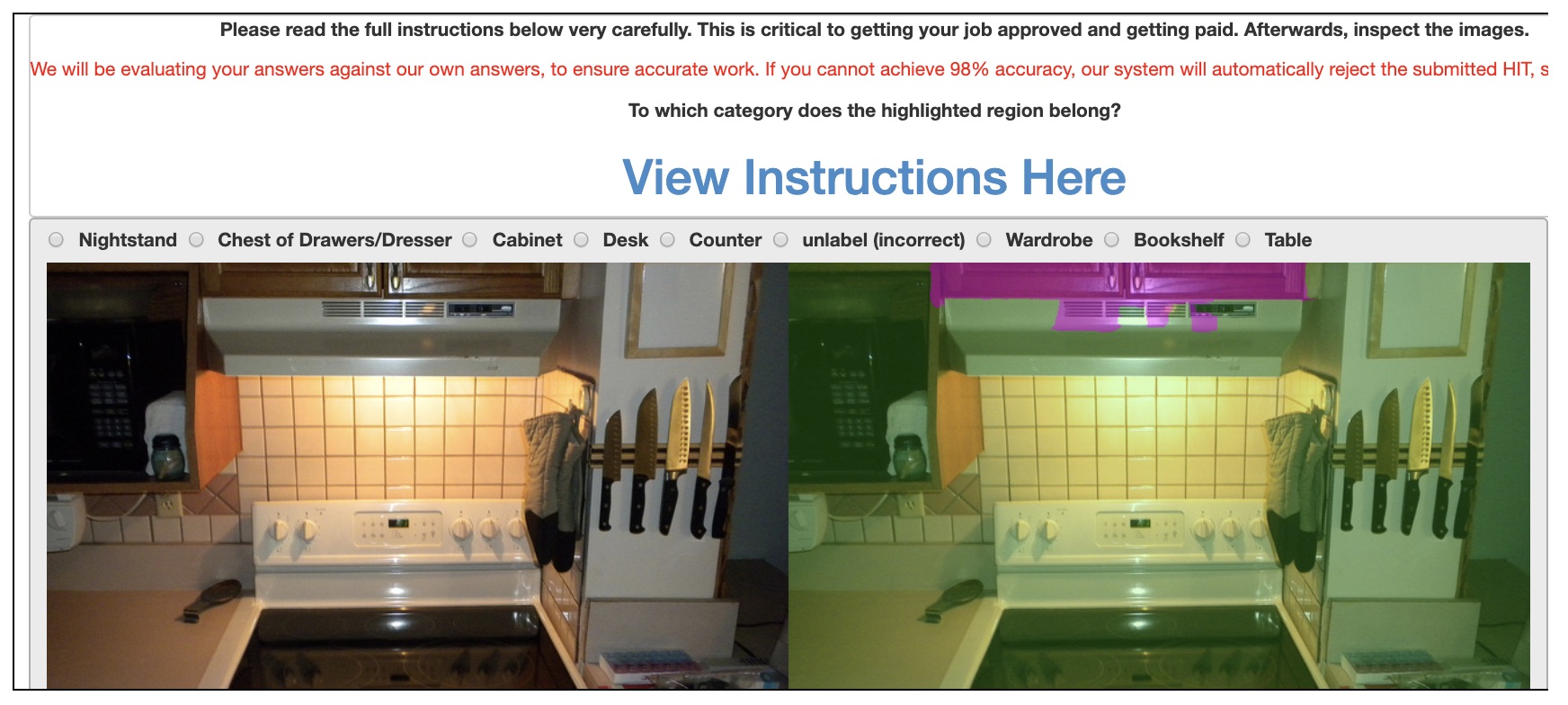}
    \includegraphics[width=0.9\columnwidth]{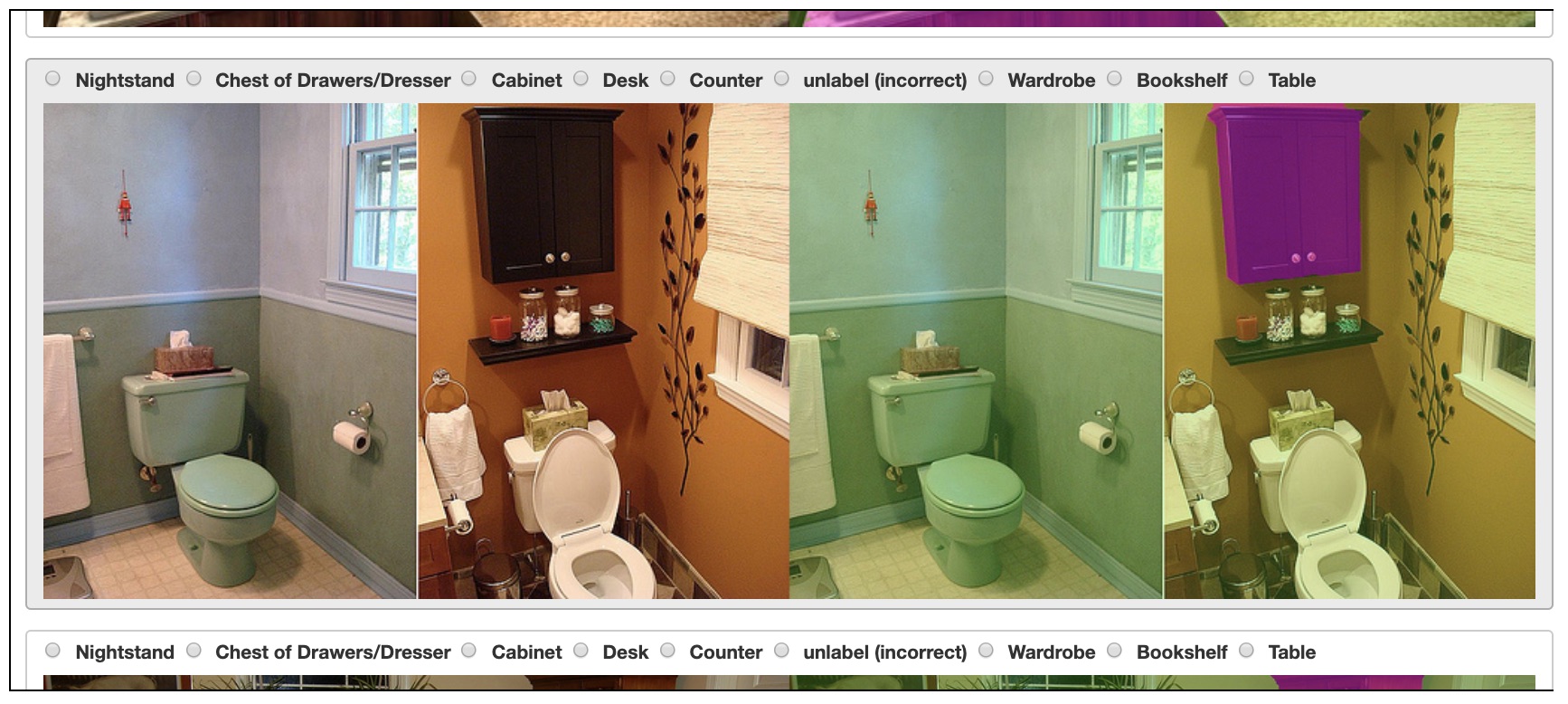}
    \includegraphics[width=0.9\columnwidth]{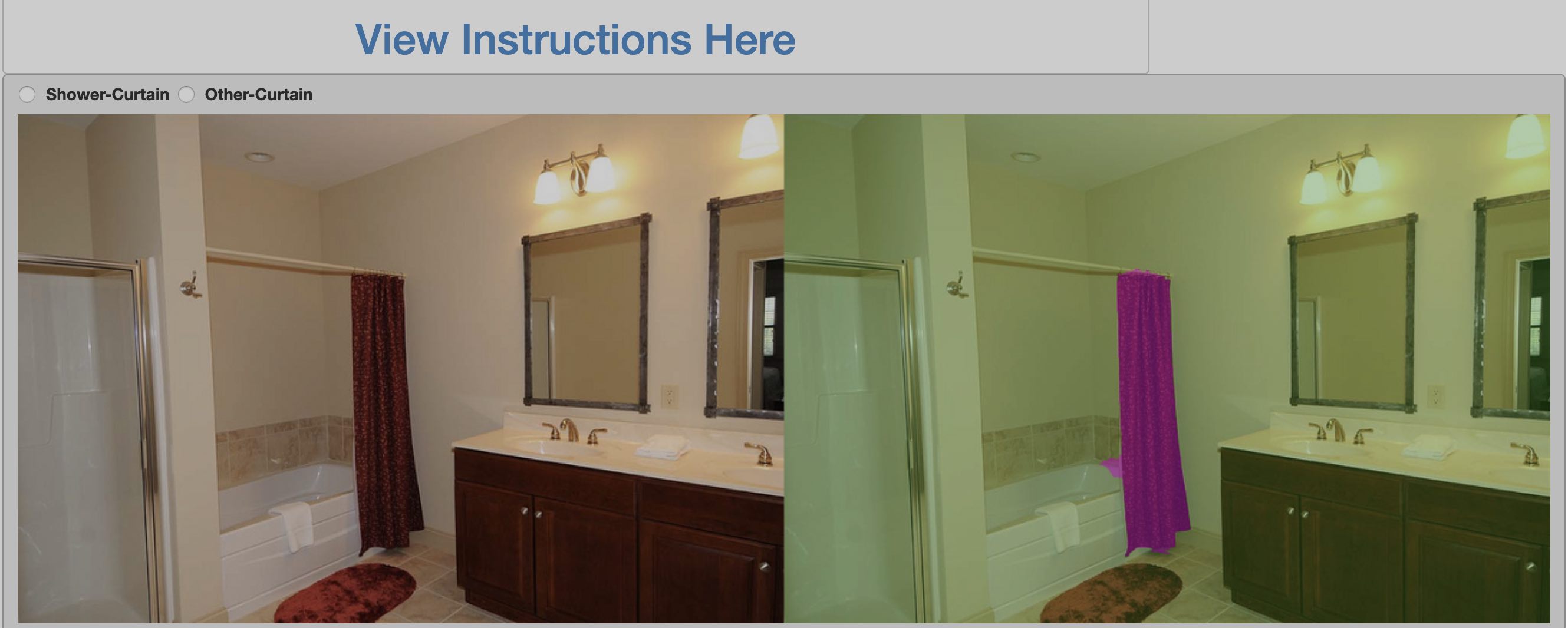}
    \includegraphics[width=0.9\columnwidth]{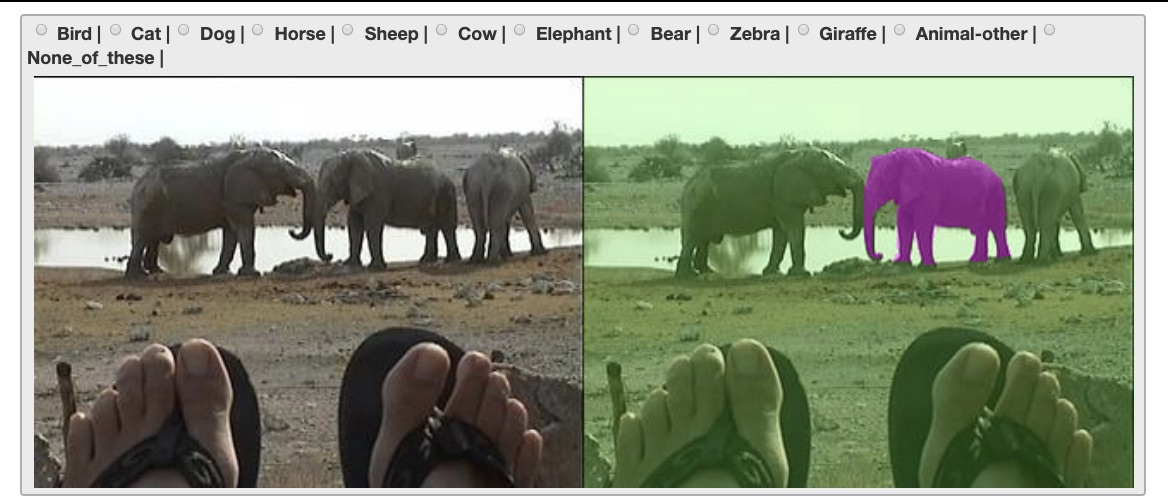}
    \caption{Example Mechanical Turk Interface. We provide binary mask classification problems to MTurk annotators. }
    \label{fig:mturkinterface}
\end{figure}

In Figure \ref{fig:mturkinterface}, we provide a sample Amazon Mechanical Turk interface for three tasks: \textit{cabinet-merged} mask splitting, \textit{curtain} mask splitting, and \textit{animal} mask splitting. We distribute human intelligence tasks (HITs) in batches of 100 masks at a time to each worker.

\section{Additional Details on MSeg}
\label{sec:msegadditionaldetails}
\begin{figure*}[t]
    \centering
    \vspace{-6mm}
    \includegraphics[width=\textwidth]{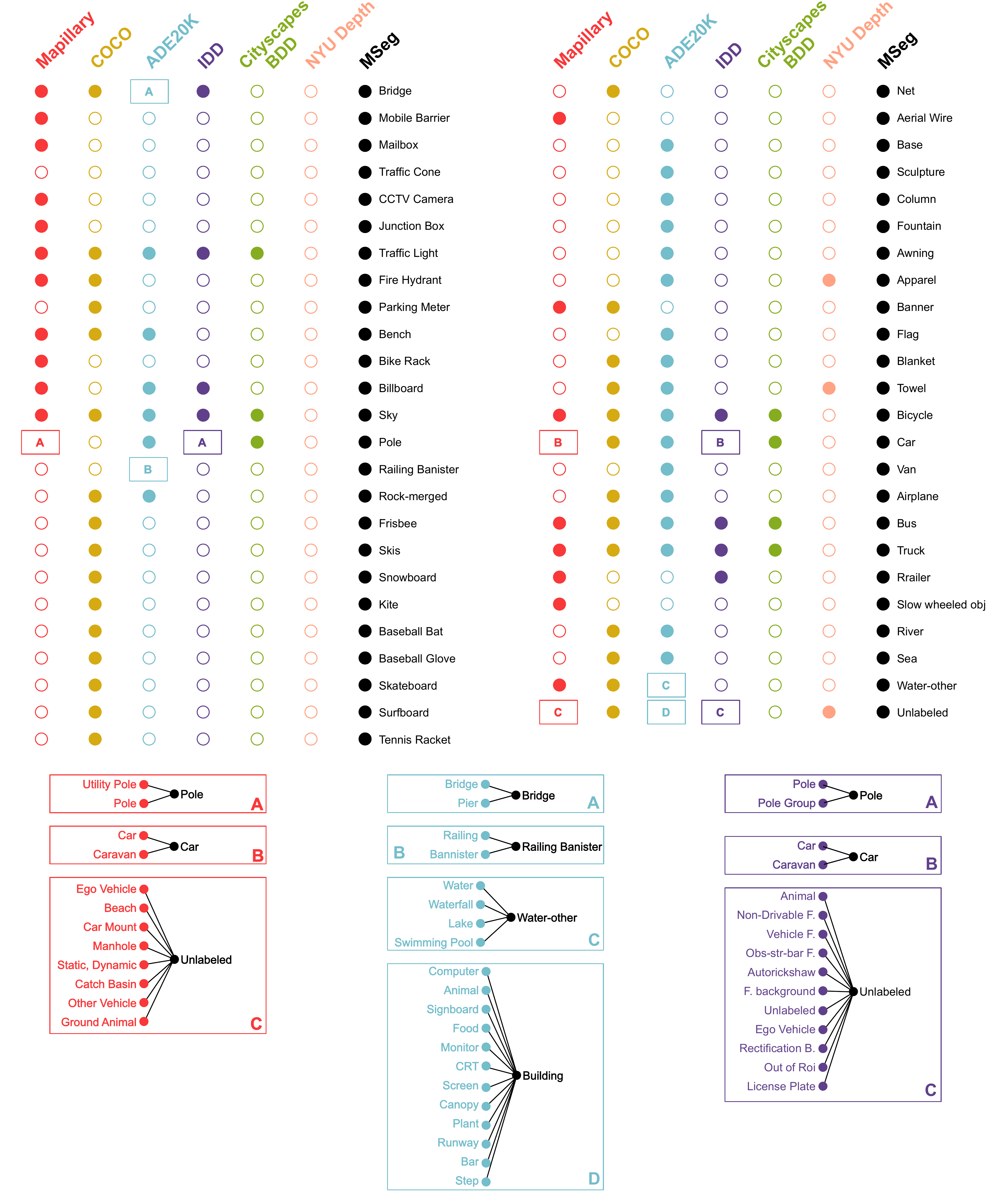}
        \vspace{-7mm}
     \caption{Visualization of a subset of the class mapping from each dataset to our unified taxonomy. This figure shows the classes omitted in the main text Figure~2. Each filled circle means that a class with that name exists in the dataset, while an empty circle means that there is no pixel from that class in the dataset. A  rectangle indicates that a split and/or merge operation was performed to map to the specified class in MSeg. Rectangles are zoomed-in in the below panel. Merge operations are shown with straight lines and split operations are shown with dashed lines. \emph{(Best seen in color.)}}
    \label{fig:alldatasettaxmapping1}
\end{figure*}

\begin{figure*}[t]
    \centering
        \vspace{-6mm}
    \includegraphics[width=\textwidth]{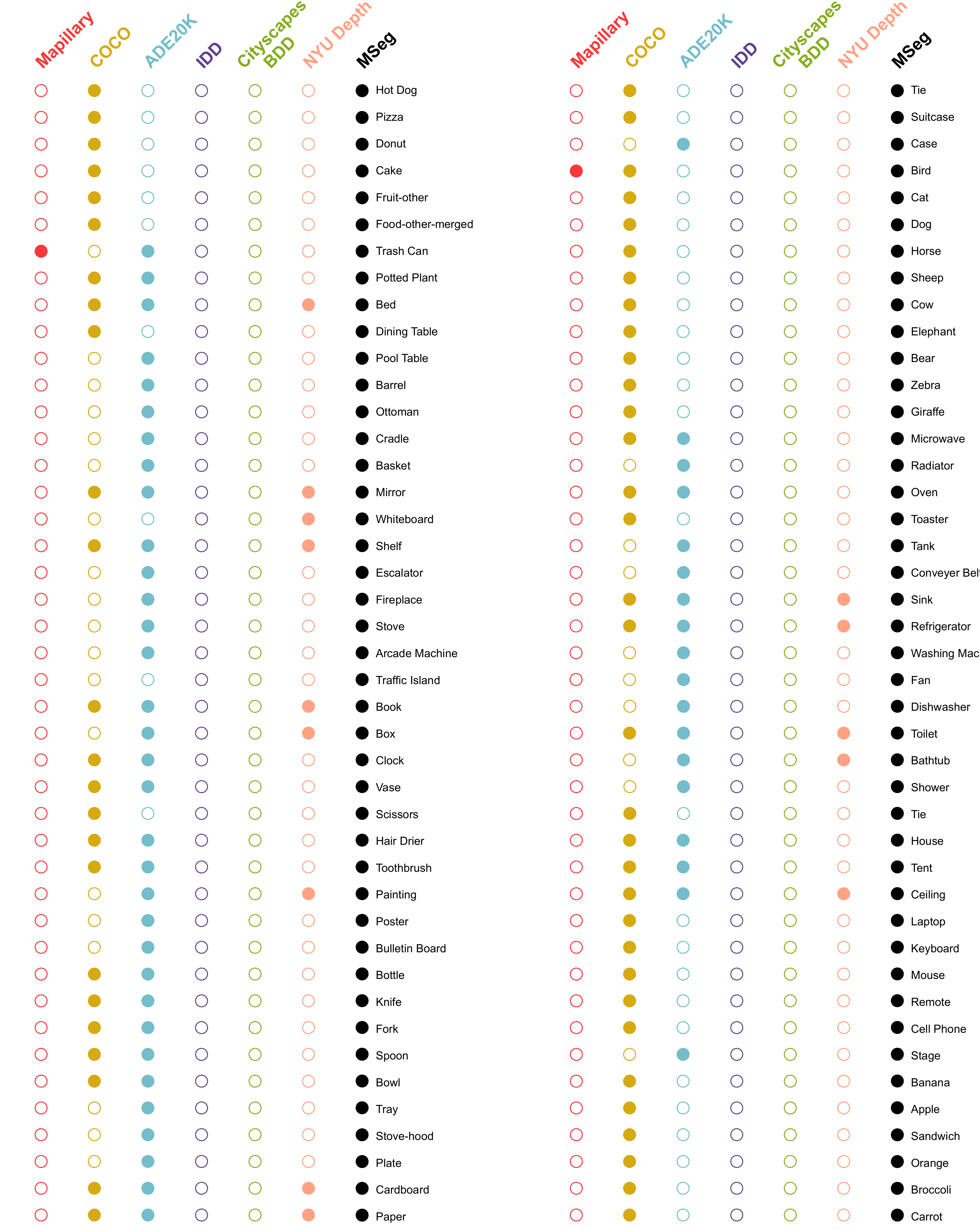}
     \caption{Visualization of a subset of the class mapping from each dataset to our unified taxonomy. This figure shows the classes omitted in the main text Figure~2. Each filled circle means that a class with that name exists in the dataset, while an empty circle means that there is no pixel from that class in the dataset. \emph{(Best seen in color.)}}
    \label{fig:alldatasettaxmapping2}
\end{figure*}

In Figure~\textcolor{blue}{2} of the main text, we visualized the mapping between the meta-training set taxonomies and the MSeg taxonomy for subset of the classes. We visualize the remaining classes in \mbox{Figures} \ref{fig:alldatasettaxmapping1} \& \ref{fig:alldatasettaxmapping2}. We also list the number of split and merge operations in Table~\ref{tab:COCO-mturk-mask-relabel-count}, \ref{tab:COCO-gz-mturk-mask-relabel-count}, \ref{tab:ADE20k-mturk-mask-relabel-count}, \ref{tab:SUNRGBD-mturk-mask-relabel-count},
\ref{tab:BDD-mturk-mask-relabel-count}, 
\ref{tab:Mapillary-mturk-mask-relabel-count}, 
\ref{tab:IDD-mturk-mask-relabel-count}, 
\ref{tab:mturk-mask-relabel-count}.

 \begin{table}[t]
\resizebox{\columnwidth}{!}{%
\begin{tabular}{@{}l@{\hspace{2mm}}l@{\hspace{1mm}}r@{}c @{\hspace{8mm}}l@{\hspace{1mm}}r@{}} 
\toprule
\multirow{2}{*}{\textbf{Dataset}} &  \multicolumn{2}{c }{\textbf{Original}} & & \multicolumn{2}{c }{\textbf{Relabeled}} \\
\cmidrule{2-3} \cmidrule{5-6}   &  
  \textbf{Class} & \textbf{Count} & & \textbf{Class} & \textbf{Count} \\
 \midrule
 \multirow{47}{*}{COCO Panoptic}   
 &  \multirow{7}{*}{bridge} &\multirow{7}{*}{1763} & & bridge & 1466 \\ 
 &  & & & building & 23 \\ 
 &  & & & unlabel & 77 \\ 
 &  & & & pier\_wharf &  194 \\ 
 &  & & & platform & 1 \\ 
 &  & & & runway & 1 \\ 
 &  & & & vegetation & 1 \\
 
 \cmidrule{2-6} &  \multirow{8}{*}{cabinet-merged} & \multirow{8}{*}{8071} & &  nightstand& 210 \\
 & &  & & chest of drawers& 106 \\
 & &  & & cabinet& 7332 \\ 
 & &  & & desk & 14 \\ 
 & &  & & counter\_other & 6 \\ 
 & &  & & unlabel (incorrect)& 371 \\
 & &  & & wardrobe & 19 \\ 
 & &  & & bookshelf& 13 \\

  \cmidrule{2-6}
 &  \multirow{10}{*}{chair} &\multirow{10}{*}{39830} & & armchair & 9205 \\ 
 &  & & & basket & 1 \\ 
 &  & & & bench & 5 \\ 
 &  & & & chair\_other & 20661 \\ 
 &  & & & unlabel & 211 \\ 
 &  & & & ottoman & 25 \\ 
 &  & & & seat & 7398 \\ 
 &  & & & slow\_wheeled\_object & 10 \\ 
 &  & & & stool & 995 \\ 
 &  & & & swivel\_chair & 1319 \\ 
 
    \cmidrule{2-6}
 &  \multirow{5}{*}{counter} &\multirow{5}{*}{4751} & & bathroom\_counter & 738 \\ 
 &  & & & counter\_other & 3823 \\ 
 &  & & & kitchen\_island & 28 \\ 
 &  & & & nightstand &  1 \\ 
 &  & & & unlabel & 161 \\ 

 \cmidrule{2-6} &   \multirow{2}{*}{curtain} &\multirow{2}{*}{5306} & & curtain\_other& 4629 \\ 
  &   & & & shower\_curtain &677 \\ 

   \cmidrule{2-6}
 &  \multirow{9}{*}{dining table} &\multirow{9}{*}{16306} & & box & 2 \\ 
 &  & & & counter\_other & 218 \\ 
 &  & & & desk & 29 \\ 
 &  & & & kitchen\_island & 135 \\ 
 &  & & & table &  15851 \\ 
 &  & & & unlabel & 68 \\ 
 &  & & & nightstand & 1 \\ 
 &  & & & chest\_of\_drawers & 1 \\ 
 &  & & & bathroom\_counter & 1 \\ 
 
 \cmidrule{2-6} &  \multirow{2}{*}{fence-merged} & \multirow{2}{*}{4856} & & fence& 4733 \\ 
 &  & & & guardrail & 123 \\ 
 
 \cmidrule{2-6} &   \multirow{6}{*}{light} & \multirow{6}{*}{12047} & &  chandelier & 260 \\
 & &  & & lamp & 4527 \\
 & &  & & light\_other & 3793 \\ 
 & &  & & unlabel (incorrect) & 86 \\ 
 & &  & & sconce & 939 \\ 
 & &  & & streetlight & 2442 \\

\bottomrule
\end{tabular}}
    \caption{We record the number of masks from each COCO Panoptic class that are relabeled (here a-f classes are shown alphabetically).
    }
    \label{tab:COCO-mturk-mask-relabel-count}
\end{table}

\begin{table}[t]
\resizebox{\columnwidth}{!}{%
\begin{tabular}{@{}l@{\hspace{2mm}}l@{\hspace{1mm}}r@{}c @{\hspace{8mm}}l@{\hspace{1mm}}r@{}} 
\toprule
\multirow{2}{*}{\textbf{Dataset}} &  \multicolumn{2}{c }{\textbf{Original}} & & \multicolumn{2}{c }{\textbf{Relabeled}} \\
\cmidrule{2-3} \cmidrule{5-6}   &  
  \textbf{Class} & \textbf{Count} & & \textbf{Class} & \textbf{Count} \\
\midrule \multirow{48}{*}{COCO Panoptic}   & \multirow{4}{*}{pavement-merged} & \multirow{4}{*}{624} & &  runway & 559 \\
 & &  & & floor & 11 \\
 & &  & & pavement\_sidewalk & 53 \\ 
 & &  & & pier\_wharf & 1 \\ 

\cmidrule{2-6} &  \multirow{7}{*}{platform} & \multirow{7}{*}{2122} & &  pier\_wharf & 236 \\
 & &  & & platform & 1616 \\
 & &  & & stage & 1 \\ 
 & &  & & playingfield & 1 \\ 
 & &  & & building & 1 \\
 & &  & & unlabel & 259 \\ 
 & &  & & bridge & 8 \\

 \cmidrule{2-6}
 & \multirow{5}{*}{person} &\multirow{5}{*}{30834} & & person-nonrider & 23819 \\ 
 &  & & & bicyclist & 2178 \\ 
 &  & & & motorcyclist & 4353 \\ 
 &  & & & rider\_other & 458 \\ 
  &  & & & unlabel & 26 \\

\cmidrule{2-6} &  \multirow{4}{*}{road} & \multirow{4}{*}{1073} & &  runway & 1023 \\
 & &  & & unlabel & 6 \\
 & &  & & pavement\_sidewalk & 12 \\ 
 & &  & & road & 32 \\ 

 \cmidrule{2-6} &  \multirow{3}{*}{rug-merged} & \multirow{3}{*}{7780} & & rug\_floormat & 3247 \\ 
 &  & & & floor (carpet) & 3993\\ 
 &  & & & unlabel (incorrect) & 540 \\

    \cmidrule{2-6}
 &  \multirow{14}{*}{table-merged} &\multirow{14}{*}{19165} & & base & 3 \\ 
 &  & & & bathroom\_counter & 77 \\ 
 &  & & & cabinet & 8 \\ 
 &  & & & chest\_of\_drawers &  1 \\ 
 &  & & & counter\_other & 637 \\ 
 &  & & & desk & 2128 \\ 
 &  & & & kitchen\_island & 67 \\
  &  & & & laptop & 1 \\ 
 &  & & & mountain\_hill & 1 \\ 
 &  & & & nightstand &  277 \\ 
 &  & & & unlabel & 7120 \\ 
 &  & & & pool\_table & 2 \\ 
 &  & & & stool & 1 \\
  &  & & & table & 8842 \\

    \cmidrule{2-6}
 &  \multirow{11}{*}{water-other} &\multirow{11}{*}{2532} & & fountain & 47 \\ 
  &  & & & unlabeled & 43 \\ 
 &  & & & playingfield & 1 \\ 
 &  & & & river\_lake & 1232 \\ 
 &  & & & sea &  81 \\ 
 &  & & & swimming\_pool & 101 \\ 
 &  & & & terrain & 1 \\ 
  &  & & & wall & 20 \\ 
 &  & & & water\_other & 985 \\
 &  & & & waterfall & 15 \\
 &  & & & window & 6 \\

\bottomrule
\end{tabular}}
    \caption{We record the number of masks from each COCO Panoptic class that are re-labeled (classes g-z, listed alphabetically.}
    \label{tab:COCO-gz-mturk-mask-relabel-count}
\end{table}

\begin{table}[t]
\resizebox{\columnwidth}{!}{%
\begin{tabular}{@{}l@{\hspace{2mm}}l@{\hspace{1mm}}r@{}c @{\hspace{8mm}}l@{\hspace{1mm}}r@{}} 
\toprule
\multirow{2}{*}{\textbf{Dataset}} &  \multicolumn{2}{c }{\textbf{Original}} & & \multicolumn{2}{c }{\textbf{Relabeled}} \\
\cmidrule{2-3} \cmidrule{5-6}   &  
  \textbf{Class} & \textbf{Count} & & \textbf{Class} & \textbf{Count} \\
 \midrule
\multirow{52}{*}{ADE20K} 
 &  \multirow{13}{*}{animal} &\multirow{13}{*}{530} & & bird & 36 \\ 
 &  & & & cat & 1 \\ 
 &  & & & dog & 2 \\ 
 &  & & & horse & 23 \\ 
 &  & & & sheep & 52 \\ 
 &  & & & cow & 115 \\ 
 &  & & & elephant & 12 \\ 
 &  & & & bear & 6 \\ 
 &  & & & zebra & 23 \\ 
 &  & & & giraffe & 10 \\ 
 &  & & & animal\_other & 243 \\ 
 &  & & & plaything & 5 \\ 
 &  & & & unlabel & 2 \\ 
\cmidrule{2-6}
 &  \multirow{5}{*}{chest\_of\_drawers} &\multirow{5}{*}{666} & & cabinet & 241 \\ 
 &  & & & chest\_of\_drawers & 403 \\ 
 &  & & & desk & 2 \\ 
 &  & & & unlabel & 15 \\ 
 &  & & & nightstand & 5 \\ 
\cmidrule{2-6}  & \multirow{2}{*}{curtain} & \multirow{2}{*}{4473} & & other-curtain & 4303 \\ 
  & & && shower-curtain & 170 \\
\cmidrule{2-6} &   \multirow{2}{*}{fence} & \multirow{2}{*}{1380} &  &  fence& 1117 \\ 
   & & & & guardrail & 263\\ 
\cmidrule{2-6}
 &  \multirow{8}{*}{food} &\multirow{8}{*}{799} & & sandwich & 10 \\ 
 &  & & & broccoli & 1 \\ 
 &  & & & pizza & 5 \\ 
 &  & & & cake & 1 \\ 
 &  & & & fruit\_other & 4 \\ 
 &  & & & food\_other & 757 \\ 
  &  & & & unlabel & 9 \\ 
 &  & & & vegetation & 12 \\ 
\cmidrule{2-6}
 &  \multirow{6}{*}{glass} &\multirow{6}{*}{1710} & & cup & 680 \\ 
 &  & & & wine\_glass & 1002 \\ 
 &  & & & window & 5 \\ 
 &  & & & unlabel & 21 \\ 
 &  & & & mirror & 1 \\ 
 &  & & & bottle & 1 \\ 
 
\cmidrule{2-6} & \multirow{2}{*}{mountain, hill} & \multirow{2}{*}{3398} & & mountain\_hill& 3313 \\ 
  & & & & snow & 85\\ 
\cmidrule{2-6}
 &  \multirow{4}{*}{person} &\multirow{4}{*}{2646} & & person-nonrider & 2415 \\ 
 &  & & & bicyclist & 127 \\ 
 &  & & & motorcyclist & 102 \\ 
 &  & & & rider\_other & 2 \\ 
\cmidrule{2-6}
 &  \multirow{2}{*}{plaything} &\multirow{2}{*}{1064} & & plaything\_other & 933 \\ 
 &  & & & teddy\_bear & 131 \\ 
\cmidrule{2-6}
&  \multirow{10}{*}{table} &\multirow{10}{*}{8017} & & table & 5988 \\ 
 &  & & & nightstand & 1716 \\ 
 &  & & & desk & 163 \\ 
 &  & & & bathroom\_counter & 1 \\ 
 &  & & & counter\_other & 24 \\ 
  &  & & & kitchen\_island & 12 \\ 
 &  & & & stool & 1 \\ 
 &  & & & desk & 49 \\ 
 &  & & & unlabel & 61 \\ 
  &  & & & cabinet & 2 \\ 
 \hline
 \bottomrule
\end{tabular}}
    \caption{We record the number of masks from each ADE20K class that are re-labeled.}
    \label{tab:ADE20k-mturk-mask-relabel-count}
\end{table}

  \begin{table}[t]
\centering
\begin{tabular}{@{}l@{\hspace{2mm}}l@{\hspace{1mm}}r@{}c @{\hspace{8mm}}l@{\hspace{1mm}}r@{}} 
\toprule
\multirow{2}{*}{\textbf{Dataset}} &  \multicolumn{2}{c }{\textbf{Original}} & & \multicolumn{2}{c }{\textbf{Relabeled}} \\
\cmidrule{2-3} \cmidrule{5-6}   &  
  \textbf{Class} & \textbf{Count} & & \textbf{Class} & \textbf{Count} \\
 \midrule
 \multirow{24}{*}{SUN RGBD}  &  \multirow{6}{*}{counter} &\multirow{6}{*}{461} & & bathroom\_counter & 78 \\ 
 &  & & & cabinet & 2 \\ 
 &  & & & counter\_other & 322 \\ 
 &  & & & desk & 3 \\ 
  &  & & & kitchen\_island & 6 \\ 
  &  & & & unlabel & 50 \\ 
\cmidrule{2-6}
&  \multirow{13}{*}{chair} &\multirow{13}{*}{3236} & & armchair & 877 \\ 
 &  & & & bench & 12 \\ 
 &  & & & cabinet & 1 \\ 
 &  & & & chair\_other & 1103 \\ 
  &  & & & door & 1 \\ 
  &  & & & unlabel & 469 \\ 
   &  & & & ottoman & 5 \\ 
 &  & & & seat & 72 \\ 
  &  & & & sofa & 5 \\ 
  &  & & & stool & 62 \\ 
  &  & & & swivel\_chair & 626 \\ 
 &  & & & table & 1 \\ 
  &  & & & wall & 2 \\ 
\cmidrule{2-6}
&  \multirow{5}{*}{lamp} &\multirow{5}{*}{775} & & chandelier & 1 \\ 
 &  & & & lamp & 708 \\ 
 &  & & & light\_other & 7 \\ 
 &  & & & unlabel & 34 \\ 
  &  & & & sconce & 25 \\ 
 \bottomrule
\end{tabular}
    \caption{We record the number of masks from each SUN RBGD class that are re-labeled.}
    \label{tab:SUNRGBD-mturk-mask-relabel-count}
\end{table}

  \begin{table}[t]
  \centering
\begin{tabular}{@{}l@{\hspace{2mm}}l@{\hspace{1mm}}r@{}c @{\hspace{8mm}}l@{\hspace{1mm}}r@{}} 
\toprule
\multirow{2}{*}{\textbf{Dataset}} &  \multicolumn{2}{c }{\textbf{Original}} & & \multicolumn{2}{c }{\textbf{Relabeled}} \\
\cmidrule{2-3} \cmidrule{5-6}   &  
  \textbf{Class} & \textbf{Count} & & \textbf{Class} & \textbf{Count} \\
 \midrule
 \multirow{9}{*}{BDD}  &  \multirow{4}{*}{person} &\multirow{4}{*}{2808} & & motorcyclist & 3 \\ 
 &  & & & bicyclist & 4 \\ 
 &  & & & person\_nonrider & 2740 \\ 
  &  & & & unlabel & 61 \\ 
\cmidrule{2-6}
 &  \multirow{5}{*}{rider} &\multirow{5}{*}{389} & & motorcyclist & 120 \\
 &  & & & bicyclist & 233 \\ 
 &  & & & person\_nonrider & 24 \\ 
  &  & & & rider\_other & 8 \\ 
 &  & & & unlabel & 4 \\ 
 \bottomrule
\end{tabular}
    \caption{We record the number of masks from each BDD class that are re-labeled.}
    \label{tab:BDD-mturk-mask-relabel-count}
\end{table}

   \begin{table}[t]
\centering
\begin{tabular}{@{}l@{\hspace{2mm}}l@{\hspace{1mm}}r@{}c @{\hspace{8mm}}l@{\hspace{1mm}}r@{}} 
\toprule
\multirow{2}{*}{\textbf{Dataset}} &  \multicolumn{2}{c }{\textbf{Original}} & & \multicolumn{2}{c }{\textbf{Relabeled}} \\
\cmidrule{2-3} \cmidrule{5-6}   &  
  \textbf{Class} & \textbf{Count} & & \textbf{Class} & \textbf{Count} \\
 \midrule
 \multirow{11}{*}{Mapillary}  &  \multirow{5}{*}{Water} &\multirow{5}{*}{500} & & fountain & 18 \\ 
 &  & & & river\_lake & 343 \\ 
 &  & & & sea & 44 \\ 
  &  & & & water\_other & 92 \\ 
  &  & & & unlabel & 3 \\ 
  \cmidrule{2-6}
 &  \multirow{6}{*}{Ground Animal} &\multirow{6}{*}{411}  & & animal\_other & 9 \\ 
 &  & & & bird & 3 \\ 
  &  & & & cat & 2 \\ 
 &  & & & cow & 3 \\ 
  &  & & & dog & 308 \\ 
   &  & & & horse & 26 \\ 
    &  & & & unlabeled & 60 \\ 
 \bottomrule
\end{tabular}
    \caption{We record the number of masks from each Mapillary Vistas class that are re-labeled.}
    \label{tab:Mapillary-mturk-mask-relabel-count}
\end{table}

   \begin{table}[t]
   \centering
\begin{tabular}{@{}l@{\hspace{2mm}}l@{\hspace{1mm}}r@{}c @{\hspace{8mm}}l@{\hspace{1mm}}r@{}} 
\toprule
\multirow{2}{*}{\textbf{Dataset}} &  \multicolumn{2}{c }{\textbf{Original}} & & \multicolumn{2}{c }{\textbf{Relabeled}} \\
\cmidrule{2-3} \cmidrule{5-6}   &  
  \textbf{Class} & \textbf{Count} & & \textbf{Class} & \textbf{Count} \\
 \midrule
 \multirow{11}{*}{IDD}   &  \multirow{11}{*}{rider} &\multirow{11}{*}{32080} & & motorcyclist & 31020 \\
 &  & & & bicyclist & 286 \\ 
 &  & & & person\_nonrider & 527 \\ 
  &  & & & rider\_other & 100 \\ 
 &  & & & unlabel & 134 \\ 
  &  & & & pole & 1 \\ 
  &  & & & backpack & 7 \\ 
  &  & & & bag & 1 \\ 
  &  & & & motorcycle & 1 \\ 
  &  & & & box & 2 \\ 
  &  & & & bicycle & 1 \\ 
 \bottomrule
\end{tabular}
    \caption{We record the number of masks from each IDD class that are re-labeled.}
    \label{tab:IDD-mturk-mask-relabel-count}
\end{table}

   \begin{table}[t]
\centering
\begin{tabular}{@{}l@{\hspace{2mm}}l@{\hspace{1mm}}r@{}c @{\hspace{8mm}}l@{\hspace{1mm}}r@{}} 
\toprule
\multirow{2}{*}{\textbf{Dataset}} &  \multicolumn{2}{c }{\textbf{Original}} & & \multicolumn{2}{c }{\textbf{Relabeled}} \\
\cmidrule{2-3} \cmidrule{5-6}   &  
  \textbf{Class} & \textbf{Count} & & \textbf{Class} & \textbf{Count} \\
 \midrule
 \multirow{6}{*}{Cityscapes}   &  \multirow{6}{*}{rider} &\multirow{6}{*}{2363} & & motorcyclist & 268 \\
 &  & & & bicyclist & 2010 \\ 
 &  & & & person\_nonrider & 25 \\ 
  &  & & & rider\_other & 48 \\ 
 &  & & & unlabel & 6 \\ 
  &  & & & bicycle & 6 \\ 
 \bottomrule
\end{tabular}
    \caption{We record the number of masks from each Cityscapes class that are re-labeled.}
    \label{tab:mturk-mask-relabel-count}
\end{table}

\section{Additional Qualitative Comparison with SwiftNet}
\begin{figure*}[]
\centering
\begin{tabular}{@{}l@{\hspace{12mm}}r@{}}
\begin{tabular}{@{\hspace{5mm}}c@{\hspace{12mm}}c@{\hspace{7mm}}c@{}}
\multicolumn{3}{@{}c@{}}{\includegraphics[width=0.45\linewidth]{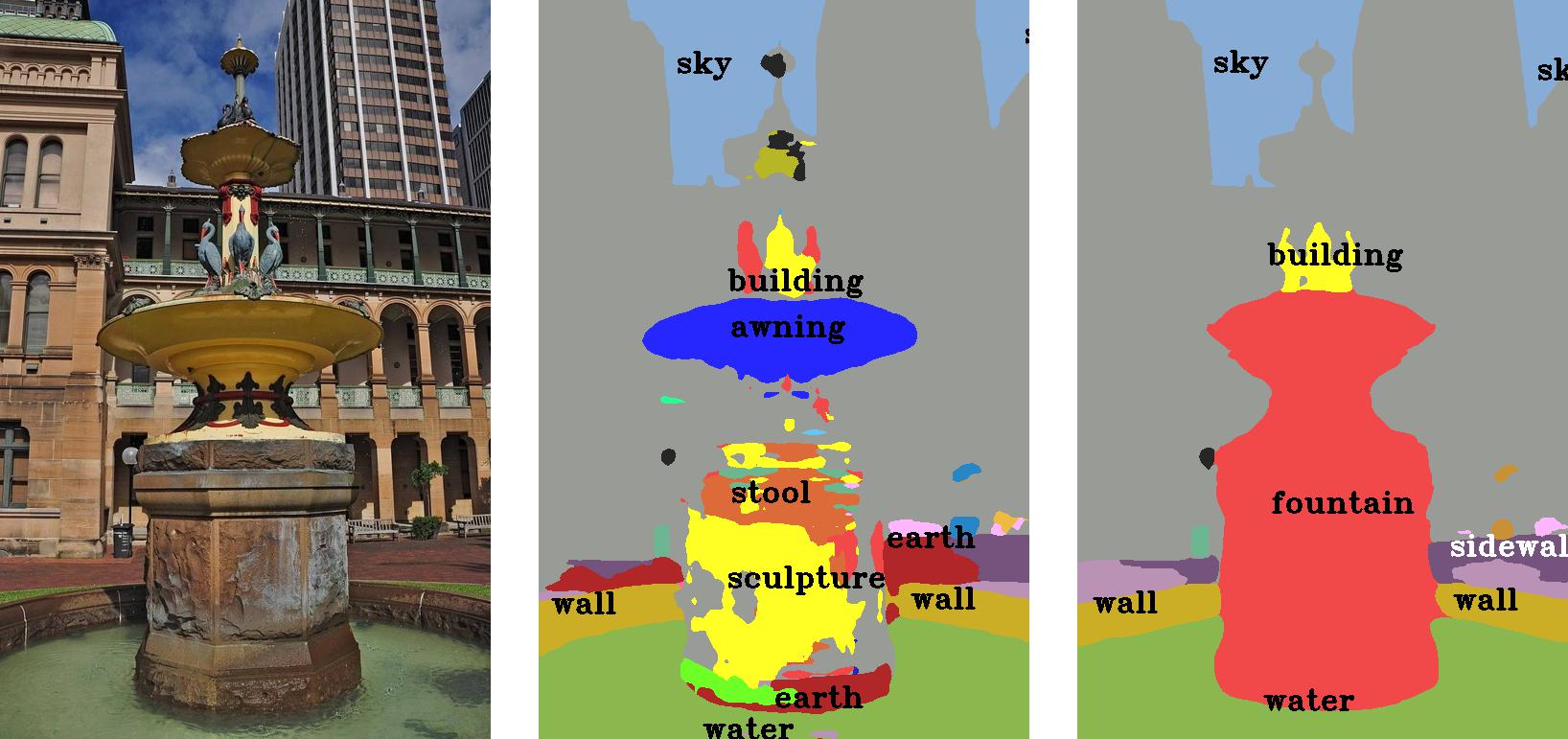}} \\
\multicolumn{3}{@{}c@{}}{\includegraphics[width=0.45\linewidth]{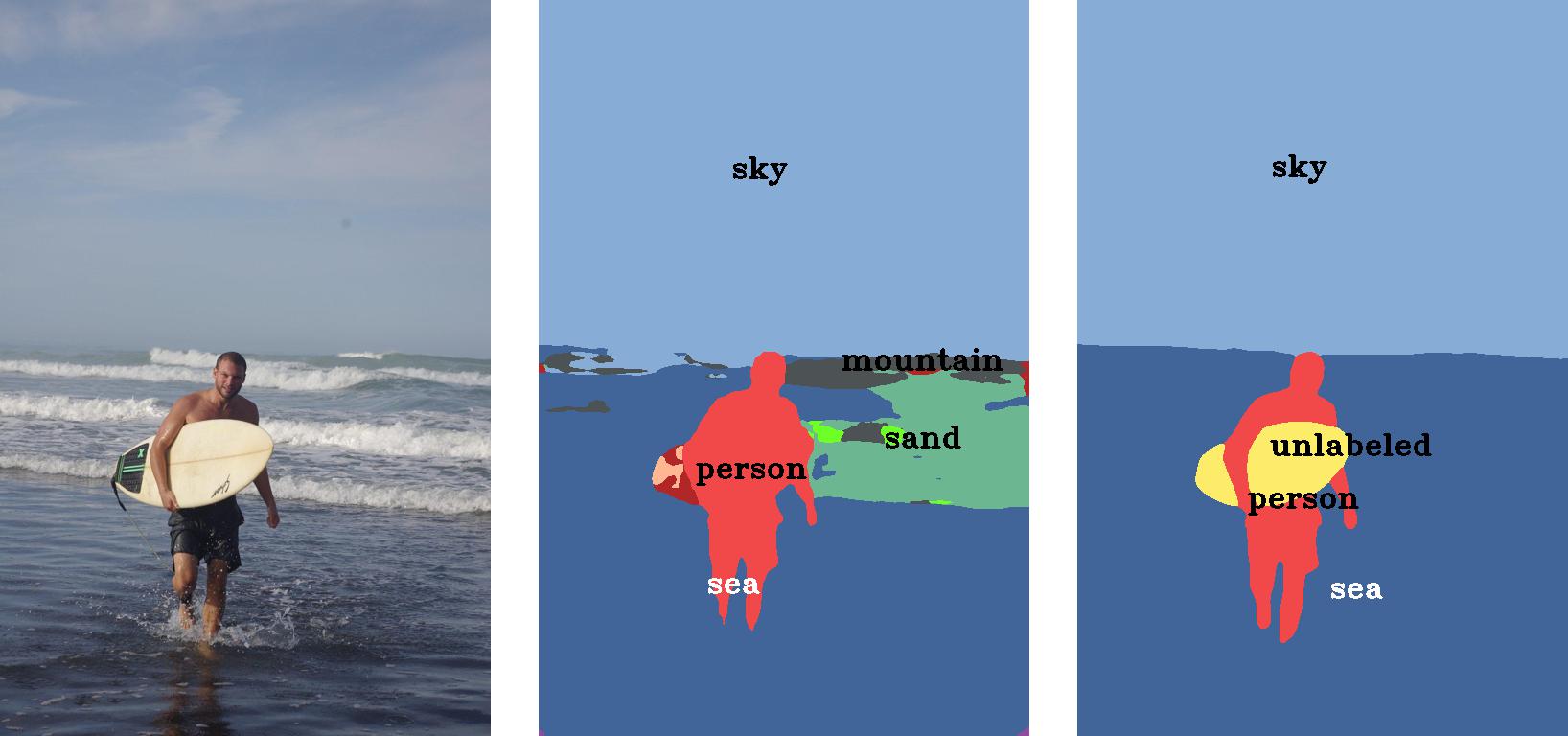}} \\
\multicolumn{3}{@{}c@{}}{\includegraphics[width=0.45\linewidth]{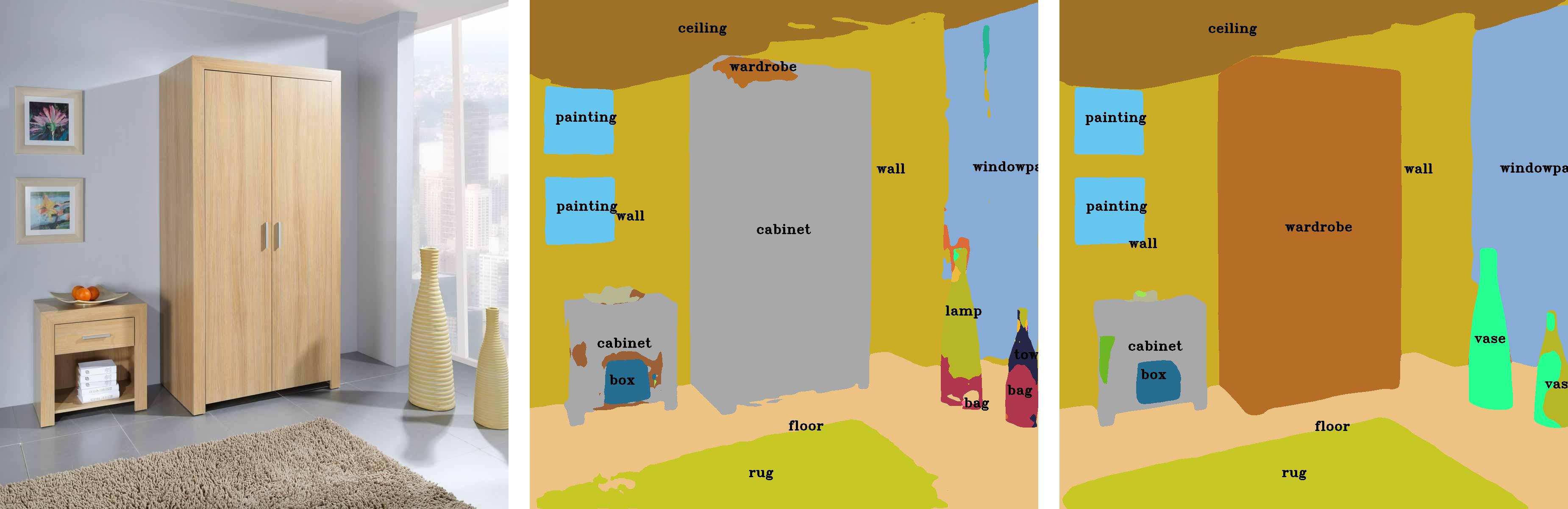}} \\
\multicolumn{3}{@{}c@{}}{\includegraphics[width=0.45\linewidth]{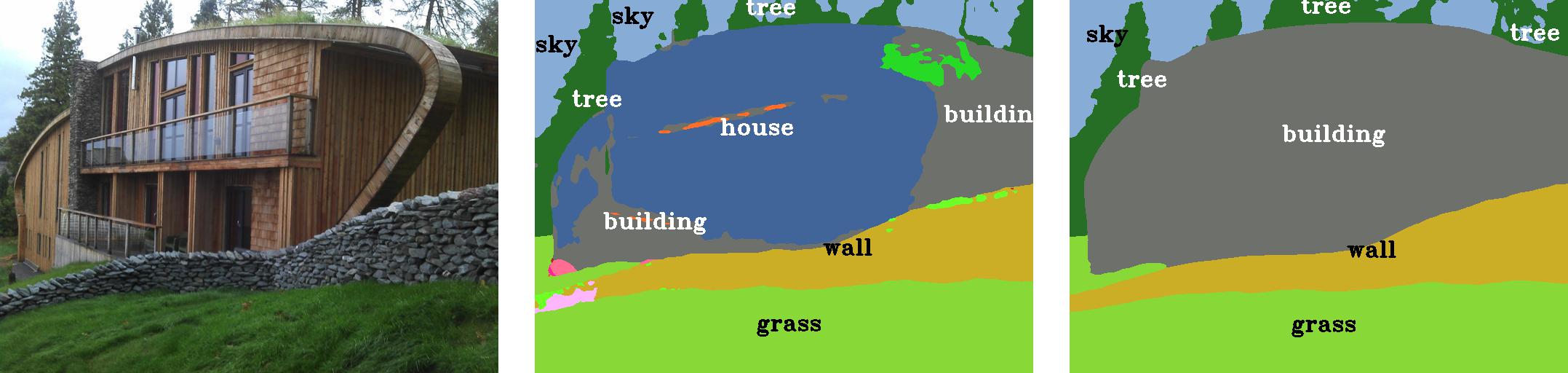}} \\
\multicolumn{3}{@{}c@{}}{\includegraphics[width=0.45\linewidth]{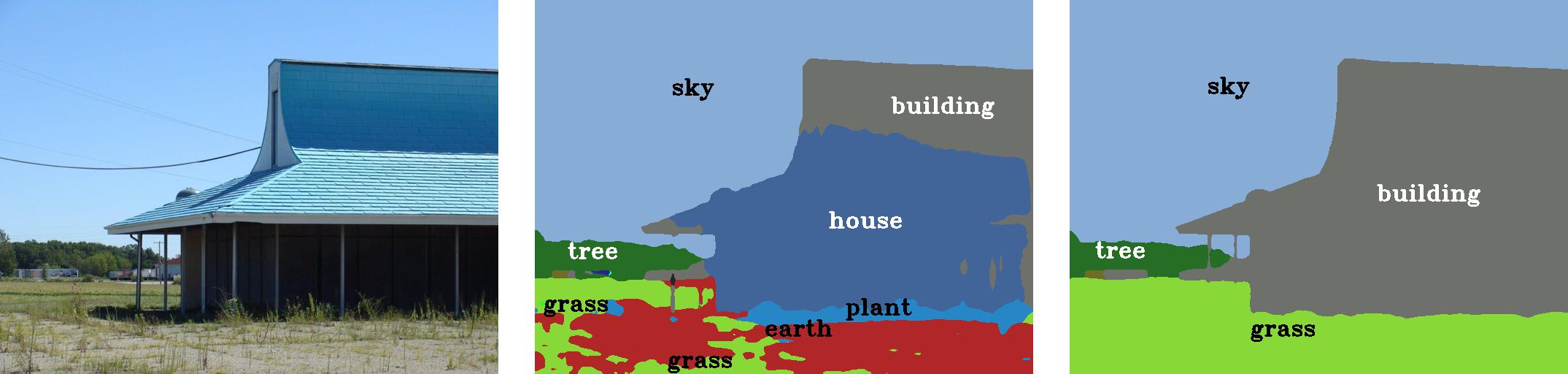}} \\
Input Image & SwiftNet & MSeg
\end{tabular} &
\begin{tabular}{@{\hspace{5mm}}c@{\hspace{12mm}}c@{\hspace{7mm}}c@{}}
\multicolumn{3}{@{}c@{}}{\includegraphics[width=0.45\linewidth]{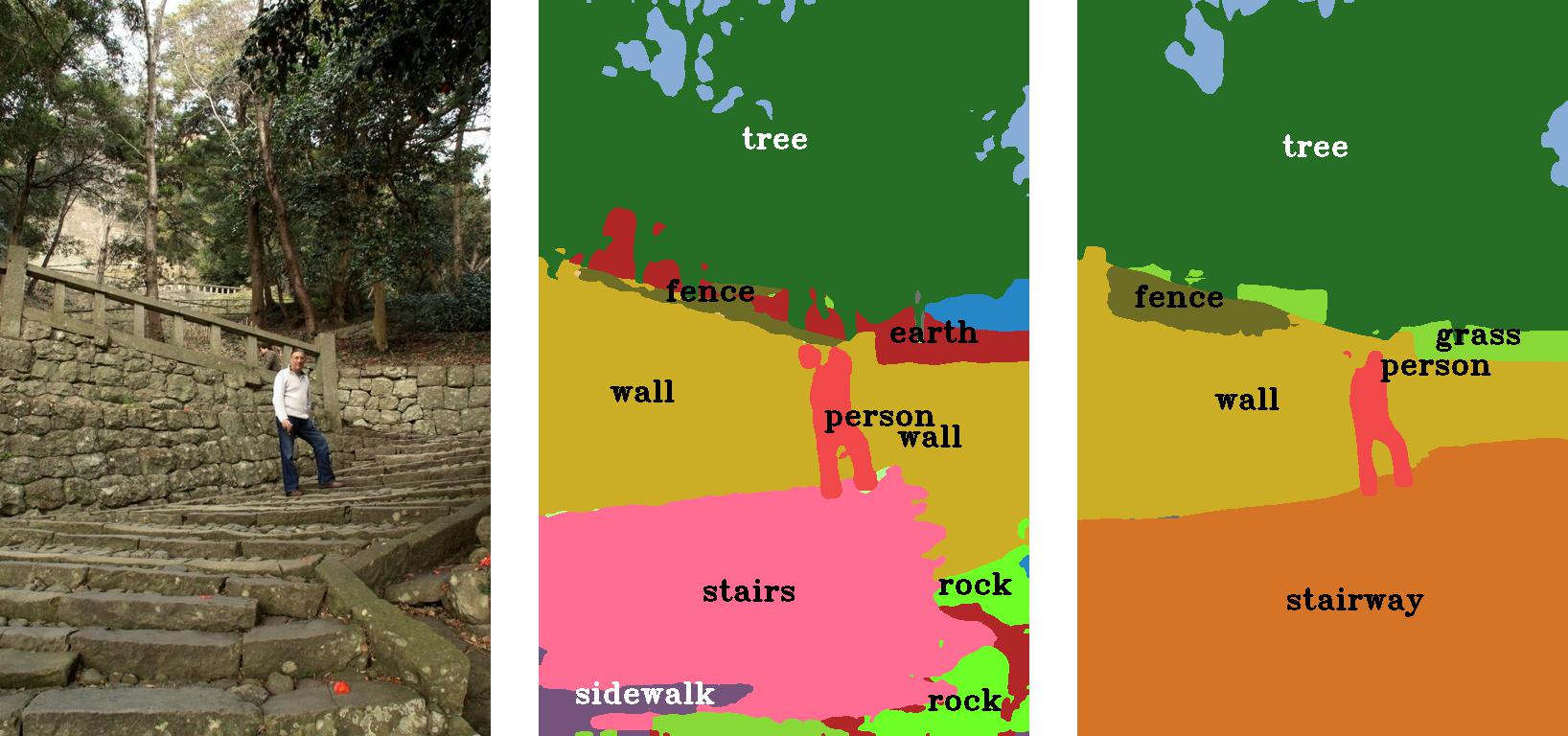}} \\
\multicolumn{3}{@{}c@{}}{\includegraphics[width=0.45\linewidth]{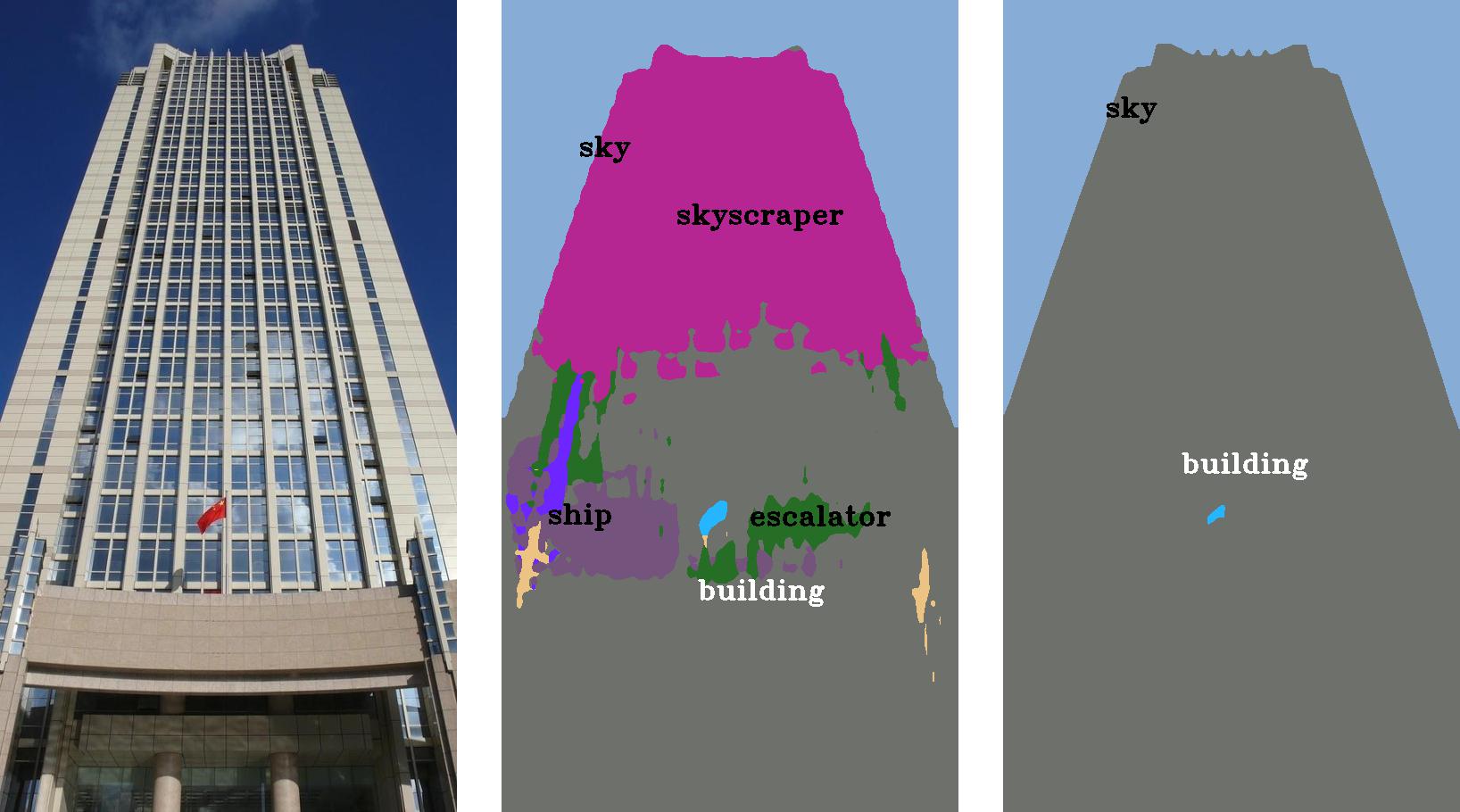}} \\
\multicolumn{3}{@{}c@{}}{\includegraphics[width=0.45\linewidth]{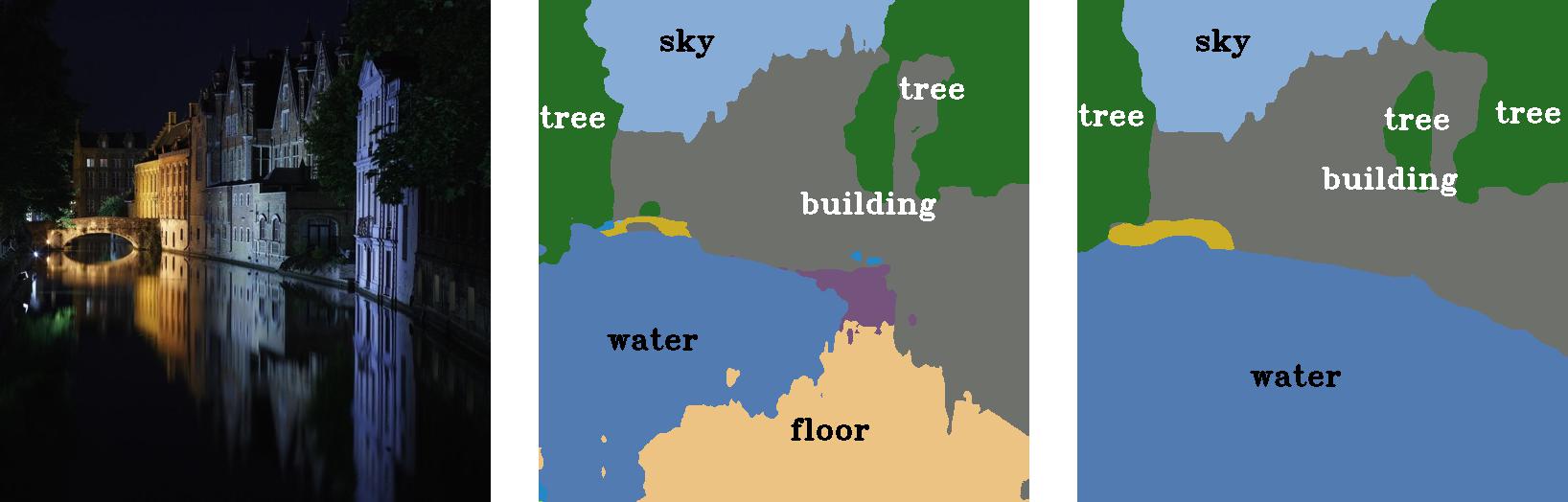}} \\
\multicolumn{3}{@{}c@{}}{\includegraphics[width=0.45\linewidth]{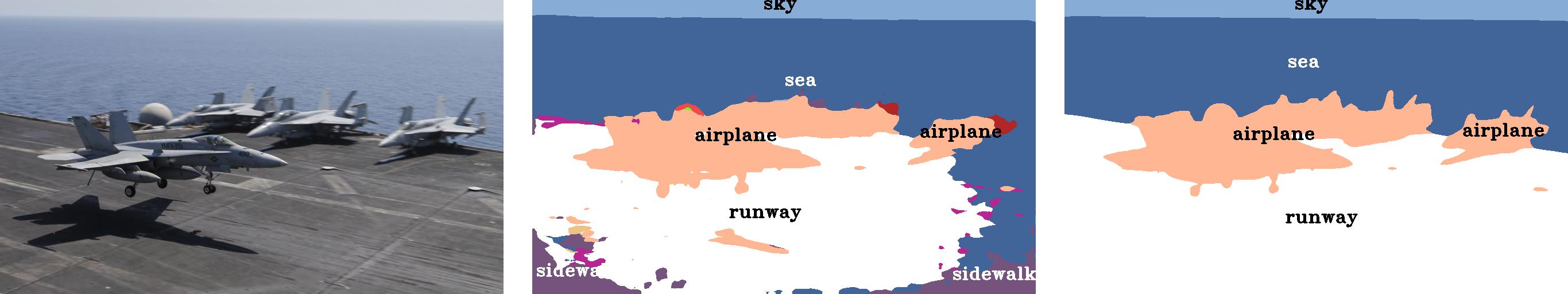}} \\
\multicolumn{3}{@{}c@{}}{\includegraphics[width=0.45\linewidth]{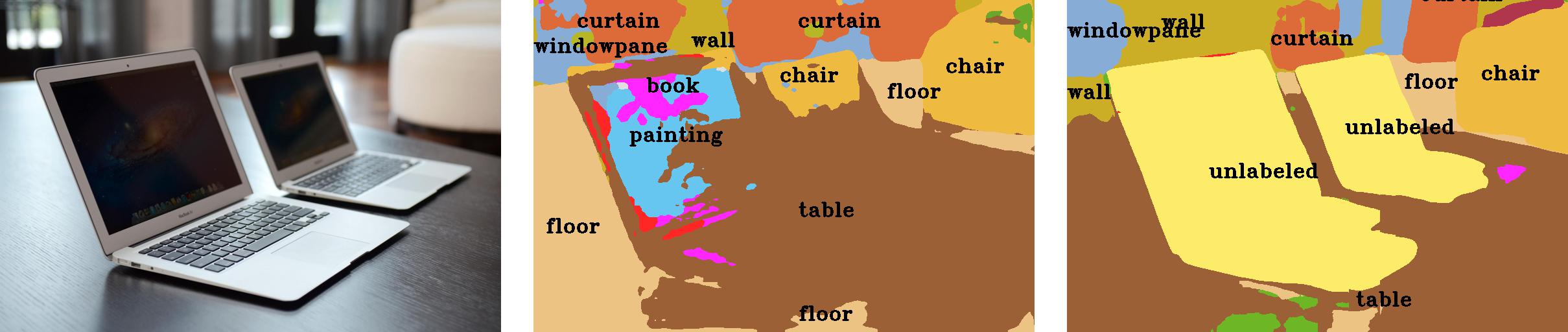}} \\
Input Image & SwiftNet & MSeg
\end{tabular}\\
\end{tabular}
 \caption{Qualitative results from SwiftNet (RVC 2020 champion) and our MSeg model (RVC 2020 runner-up without any re-training on RVC datasets) on images from the \textbf{ADE20K} RVC test dataset (test split). Once again taxonomies greatly influence predictions; We found that Mechanical Turk workers were unable to consistently separate skyscrapers, buildings, and house instances, therefore our taxonomy merges them and our model avoids confusion between the three. }
 \label{fig:ade20k}
\end{figure*}

In addition to the qualitative study on WildDash-v2 in the main paper, we qualitatively compare the segmentation performance of our method with SwiftNet on ADE20K in Figure~\ref{fig:ade20k}.

\section{Panoptic Class Mapping}

\noindent \textbf{Resolving Stuff-Thing Inconsistencies.} As discussed in the main text, each of our training datasets has its own set of `stuff' and `thing' classes and these inconsistencies must be resolved for unified training. This occurs because the delineation between `stuff' and `thing' classes can be ambiguous as times.

In Table \ref{tab:panoptic-stuff-thing-decisions}, we show the stuff vs. thing assignments in MSeg. As also discussed in the main text, to differentiate between ``thing'' and ``stuff'' classes in the universal taxonomy, we deem the class to be a ``thing'' if all classes mapped to that universal class are ``thing'' classes in their original datasets. If a class is considered `stuff' in even a single dataset, this pollutes the class, and must be considered `stuff` for all datasets.

% This step is necessary since instance segmentation detects and segments ``thing'' objects, while panoptic segmentation will also segment ``stuff'' pixels. % TODO: resolve this unclarity with Ozan

% We evaluate these models in the datasets with instance annotations. Since most test datasets in MSeg do not have these annotations, we only evaluate with MSeg training datasets. Out of these, we exclude BDD and SUN RGBD because they do not provide instance labels. We use validation sets of these datasets in our evaluation. 

\begin{table*}[]
\caption{The classification into `stuff` and `things' by origin datasets creates inconsistencies. Below, `thing' classes are shown in \textcolor{thing}{red}, and `stuff' is shown in \textcolor{stuff}{blue}. BDD and SUN RGBD do not provide instance labels.}
\begin{adjustbox}{max width=2\columnwidth}
\begingroup
\renewcommand{\arraystretch}{1.25}
\begin{tabular}{llllll}
\toprule
Unified Class Name & ADE20K                                                             & Cityscapes                               & COCO-Panoptic                                                               & IDD                                                       & Mapillary  \\
\midrule 
building &	\textcolor{thing}{booth}, \textcolor{stuff}{house, building} &	\textcolor{stuff}{building} &	\textcolor{stuff}{building-other-merged, house, roof}&	\textcolor{stuff}{building} &	\textcolor{thing}{Phone Booth}, \textcolor{stuff}{Building} \\
& \textcolor{stuff}{skyscraper, hovel, tower, grandstand} & & & & \\ 
food\_other      & \textcolor{thing}{food}                                     &                                          & \textcolor{stuff}{food-other-merged}                                  &                                                           &                                                                                                                                                                 \\
table            & \textcolor{thing}{coffee table, table}                        &                                          & \textcolor{thing}{dining table}, \textcolor{stuff}{table-merged} &                                                           &                                                                                                                                                                 \\
counter\_other   & \textcolor{thing}{counter}                                   &                                          & \textcolor{stuff}{counter}                                            &                                                           &                                                                                                                                                                 \\
door             & \textcolor{thing}{door, screen door}                         &                                          & \textcolor{stuff}{door stuff}                                         &                                                           &                                                                                                                                                                 \\
light\_other     & \textcolor{thing}{light}                                     &                                          & \textcolor{stuff}{light}                                              &                                                           &                                                                                                                                                                 \\
mirror           & \textcolor{thing}{mirror}                                    &                                          & \textcolor{stuff}{mirror stuff}                                       &                                                           &                                                                                                                                                                 \\
shelf            & \textcolor{thing}{shelf}                                     &                                          & \textcolor{stuff}{shelf}                                              &                                                           &                                                                                                                                                                 \\
stairs           & \textcolor{thing}{stairs}                                    &                                          & \textcolor{stuff}{stairway, stairs}                                   &                                                           &                                                                                                                                                                 \\
cabinet          & \textcolor{thing}{cabinet}, \textcolor{stuff}{buffet}   &                                          & \textcolor{stuff}{cabinet-merged}                                     &                                                           &                                                                                                                                                                 \\
\hline
road             & \textcolor{stuff}{road}                                      & \textcolor{stuff}{road}            & \textcolor{stuff}{road}                                               &  \textcolor{stuff}{road, parking, drivable fallback} & \textcolor{thing}{Crosswalk-Plain, Lane Marking-Crosswalk} \\
& & & & & \textcolor{stuff}{Pothole, Parking, Road, Bike Lane, } \\
& & & & & \textcolor{stuff}{Service Lane, Lane Marking-General} \\
\hline
box              & \textcolor{thing}{box}                                       &                                          & \textcolor{stuff}{cardboard}                                          &                                                           &                                                          \\
\hline
traffic\_sign    &                                                                    & \textcolor{stuff}{traffic sign}    & \textcolor{thing}{stop sign}                                          & \textcolor{stuff}{traffic sign}                     &  \textcolor{thing}{Traffic Sign (Back), Traffic Sign (Front)}       \\
& & & & &  \textcolor{thing}{Traffic Sign Frame} \\
\hline
traffic\_light   & \textcolor{thing}{traffic light}                             & \textcolor{stuff}{traffic light}   & \textcolor{thing}{traffic light}                                      & \textcolor{stuff}{traffic light}                    & \textcolor{thing}{Traffic Light}                                                                                                                          \\
billboard        & \textcolor{thing}{trade name}                                &                                          &                                                                             & \textcolor{stuff}{billboard}                        & \textcolor{thing}{Billboard}                                                                                                                              \\
\hline
pole             & \textcolor{thing}{pole}                                      & \textcolor{stuff}{pole}            &                                                                             & \textcolor{stuff}{pole, polegroup}                  & \textcolor{thing}{Utility Pole,Pole}                                                                                                                      \\
\hline
fence            & \textcolor{thing}{fence}                                     & \textcolor{stuff}{fence}           & \textcolor{stuff}{fence-merged}                                       & \textcolor{stuff}{fence}                            & \textcolor{stuff}{Fence}   \\
\hline
banner           &                                                                    &                                          & \textcolor{stuff}{banner}                                             &                                                           & \textcolor{thing}{Banner}     \\
\hline
curtain\_other   & \textcolor{thing}{curtain}                                   &                                          & \textcolor{stuff}{curtain}                                            &                                                           & \\
\hline
pillow           &                                                                    & \textcolor{thing}{pillow, cushion} & \textcolor{stuff}{pillow}                                             &                                                           &    \\
\hline
towel            & \textcolor{thing}{towel}'                                    &                                          & \textcolor{stuff}{towel}'                                             &                                                           & \\
\hline
vegetation       & \textcolor{thing}{flower,palm} \textcolor{stuff}{tree} & \textcolor{stuff}{vegetation}      & \textcolor{stuff}{flower, tree-merged}                                & \textcolor{stuff}{vegetation}                       & \textcolor{stuff}{Vegetation}                                \\
\hline
train            &                                                                    & \textcolor{thing}{train}           & \textcolor{thing}{train}                                              & \textcolor{thing}{train}                            & \textcolor{stuff}{On Rails}  \\
\hline
window           & \textcolor{thing}{windowpane}                                &                                          & \textcolor{stuff}{window-other}                                       &             &       \\                                    \bottomrule       
\end{tabular}
\endgroup
\end{adjustbox}
\label{tab:panoptic-stuff-thing-decisions}
\end{table*}

\section{Additional Re-labeling for Denoising}
\begin{figure}[ht]
    \centering
    \includegraphics[width=\columnwidth]{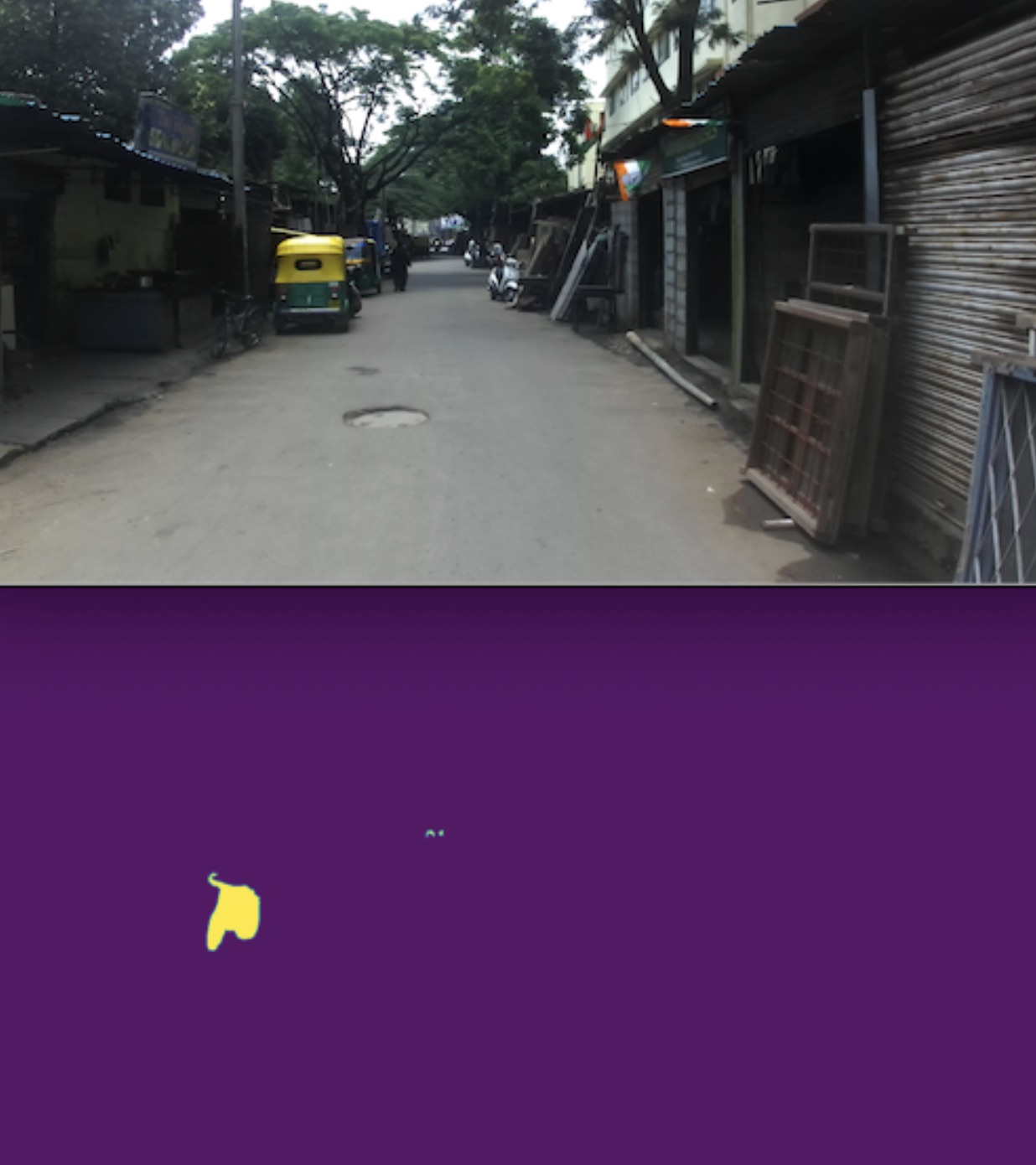}
    \caption{An example error from IDD\cite{Varma:WACV19:IDD} segmentation dataset. The mask above is labeled as class 8 (“rider”), which we relabel to class 10 (“bicycle”)}
    \label{fig:example_error}
\end{figure}

In our qualitative analysis of the datasets, we observed abundance of erroneous annotations for some specific classes in IDD\cite{Varma:WACV19:IDD}. We added such an example error in Figure~\ref{fig:example_error} to visualize the issue. Hence, to solve these errors, we performed additional re-labelling for the IDD for rider, bicycle, and motorcycle classes.

%\vfill


\end{document}